\documentclass{article} % For LaTeX2e
\usepackage{iclr2027_conference,times}

\usepackage{amsmath,amsfonts,bm}

\def\eqref#1{equation~\ref{#1}}
\def\1{\bm{1}}

\DeclareMathAlphabet{\mathsfit}{\encodingdefault}{\sfdefault}{m}{sl}
\SetMathAlphabet{\mathsfit}{bold}{\encodingdefault}{\sfdefault}{bx}{n}

\usepackage[table]{xcolor}      % load before packages that use xcolor
\usepackage{amsmath,amssymb}    % equation*, \text, \tfrac, \mathbb
\usepackage{graphicx}
\usepackage{booktabs}           % \toprule, \midrule, \bottomrule
\usepackage{array}
\usepackage{tabularx}
\usepackage{longtable}
\usepackage{wrapfig}
\usepackage{kotex}              % Korean; compile with XeLaTeX or LuaLaTeX
\usepackage{hyperref}           % keep near the end
\usepackage{xurl}               % load after hyperref; also loads url
\usepackage{tcolorbox}
\usepackage{fvextra}
\usepackage{placeins}
\usepackage{tcolorbox}
\tcbuselibrary{breakable}
\usepackage{multirow}
\usepackage{placeins}
\usepackage{booktabs}
\usepackage{graphicx}
\usepackage{caption}

\definecolor{SyntheticPromptBackground}{RGB}{245,240,250}
\definecolor{SyntheticPromptPurple}{RGB}{112,48,160}

\newsavebox{\fakeSourceBox}
\newsavebox{\realSourceBox}

\newcolumntype{Y}{>{\raggedright\arraybackslash}X}
\newcolumntype{N}{>{\raggedleft\arraybackslash}p{0.85cm}}

\newcommand{\corig}{\textsc{Original}}
\newcommand{\chid}{\textsc{Hidden}}
\newcommand{\cred}{\textsc{URL-only}}

\newcommand{\srcpos}{\textsc{Source}$+$}
\newcommand{\srcneg}{\textsc{Source}$-$}
\newcommand{\srczero}{\textsc{Source}$0$}

\title{Source Preference in the Wild: \\How LLM Agents Favor Items by Source, \\and How to Reduce It}

\author{Jonghyun Song\thanks{\ \ Equal contribution.}\hspace{1.5em}
Haewon Park\footnotemark[1]\hspace{1.5em}
Jeonghoon Shim\footnotemark[1]\hspace{1.5em}
Woojung Song\hspace{1.5em}
Yohan Jo\thanks{\ \ Corresponding author.} \\
{\mdseries Graduate School of Data Science, Seoul National University}\\
{\mdseries\ttfamily \{hyeongoon11, dellaanima2, jhshim98, yohan.jo\}@snu.ac.kr}
}

\iclrfinalcopy % Uncomment for camera-ready version, but NOT for submission.
\begin{document}

\maketitle

\begin{abstract}
As LLM agents decide on users' behalf which product to buy, which hotel to book, or which paper to cite, a preference for items from certain sources (the sites or services they come from) shapes what users receive and which sources are selected. We study source preference in end-to-end search with 12 agent models across three domains. Comparing items from different sources that satisfy the same requirements at the same position, we find that each model prefers some sources and avoids others in every domain, largely agreeing on which. This preference can outweigh how well items satisfy the request: an item satisfying one requirement fewer is selected about two-thirds of the time when it comes from a preferred source and the better one from a dispreferred source, but almost never in the reverse case. The information identifying an item's source affects selection by itself: hiding it weakens the preference, and relabeling an item with a preferred source raises its selection rate. We test two routes to this preference: training that rewards better items can make a source a shortcut for requirement satisfaction, and missing information can trigger preconceptions about the source. Supplying missing information or a prompt countering these preconceptions reduces source preference.
\end{abstract}

\section{Introduction}

Large language model agents increasingly move beyond providing information to deciding what users see.
They decide which products to show \citep{yao2022webshop,zhou2024webarena}, which accommodations to recommend \citep{xie2024travelplanner,hadad2026hotelquest}, and which papers to present \citep{skarlinski2024language,asai2026synthesizing,shen2026scholargym}.
Consider an agent searching for accommodations from \emph{sources} such as Booking.com and Expedia.
The agent has a \emph{source preference} when it favors items from one source over another, even though they satisfy the request equally well.
Such a preference can narrow the options available to users and lead agents to pass over better items because of their source.
If these selections subsequently feed back into training, existing preferences could become stronger, concentrating visibility on favored sources while leaving others overlooked even when they offer equally good items.
Prior work provides evidence of such preferences, but has not examined this influence in real-world settings where agents retrieve and select items themselves \citep{dai2025media,khan2026agents,schuster-etal-2026-whose}.
For example, \citet{khan2026agents} present a few items with semantically identical content attached to different sources and show that models favor some sources over others.
In actual search, however, each source offers its own items, which differ in content and how well they satisfy the user's request.
We therefore study whether source preference appears in end-to-end search, whether the information identifying an item's source itself contributes to such preferences, and how this reliance on the source might arise and be reduced.
Our study covers 12 agent models and three domains: shopping, accommodation, and scholarly search.

Establishing source preference in end-to-end search requires accounting for differences in the retrieved items.
A higher selection rate alone is insufficient: a source's items may better satisfy the request or appear in more favorable positions~\citep{zheng2023judging,allouah2025aiagentbuyingevaluation}.
We therefore compare items from different sources that satisfy the same requirements, with position held fixed, to score each source's preference and classify it as preferred or dispreferred (\S\ref{sec:measurement}).
Under these controls, every model prefers some sources and avoids others in every domain, with most frequently retrieved sources falling into either group.
Models largely agree on these preferences: most models prefer Booking.com, while many avoid Expedia, even though the items from these sites satisfy the same requirements.
We next examine whether these preferences persist when the preferred source offers a less satisfying item.
To do so, we compare two items, one of which satisfies one fewer requirement than the other.
Agents select the less satisfying item about two-thirds of the time when it comes from a preferred source and the better item from a dispreferred source.
They almost never select the less satisfying item when it comes from a dispreferred source and the better item from a preferred source.
This suggests that source preference can leave users with less satisfying items and cause dispreferred sources to lose selections even when they offer better ones (\S\ref{sec:natural}).

These comparisons establish source preference under the same conditions, but do not isolate the contribution of the information identifying an item's source.
We therefore examine whether changing this information changes selection while keeping the items fixed.
First, we present the same search result lists with this information hidden and then restored: hiding it weakens source preference, and restoring it widens the gap between preferred and dispreferred sources.
We then relabel the same items with preferred and dispreferred sources, keeping their titles and content fixed: across every model and domain, selection rates are higher when the same content is labeled with a preferred source than with a dispreferred one.
These interventions show that this information itself contributes to agents' selections (\S\ref{sec:origin}).

Given that the source itself affects selection, why might agents use it as a selection signal?
We examine how this reliance can arise during training and be triggered at inference.
During training, an agent may learn to use the source as a shortcut for requirement satisfaction: when a source is more often paired with the better item during DPO (\citealp{rafailov2024directpreferenceoptimizationlanguage}), agents learn to prefer it,  while balancing this pairing can mitigate an existing preference (\S\ref{sec:dpo}).
At inference, missing information may trigger the agent's preconceptions about the source, which fill the gap; supplying the missing information or prompting the agent to counter such preconceptions reduces this reliance (\S\ref{sec:rationale-rationale}).

Together, our findings suggest that agents can use the source as a shortcut for requirement satisfaction, with consequences for which items users see.
Our experiments connect preferences observed in end-to-end search to the influence of the source itself and identify a training association that can produce such preferences.
They also show that providing the missing information and countering preconceptions about the source reduce this reliance, helping agents base their selections on how well the items themselves meet the request.

\section{End-to-End Agent Setting}
\label{sec:setting}

\paragraph{Agents and Search Workflow.}
We evaluate GPT-\{5.4-nano ~\citep{gpt5.4}, 5.6-Luna~\citep{gpt5.6}\}, Gemini-3.7-Flash~\citep{gemini3.7}, GLM-5.3-Flash ~\citep{glm5team2026glm5vibecodingagentic}, DeepSeek-v4-Flash-0731~\citep{deepseekai2026deepseekv4highlyefficientmilliontoken}, Llama-4-\{Maverick, Scout\}~\citep{llama4}, Llama-3.1-8B-Instruct~\citep{grattafiori2024llama3herdmodels}, Tulu-3-8B~\citep{lambert2025tulu3pushingfrontiers}, Qwen3.5-27B~\citep{qwen3.5}, Qwen3-30B-A3B~\citep{yang2025qwen3technicalreport}, and Qwen2.5-32B-Instruct~\citep{qwen2025qwen25technicalreport} as agents in an end-to-end search-and-selection setting. 
All agents follow a ReAct-style interaction loop \citep{yao2022react}, alternating reasoning and actions with observations from the search environment under a shared prompt.
Given a user request, the agent issues a search query and receives up to 10 results, each containing a title, content, and URL. 
The agent then evaluates these results against the request and may select multiple items or issue another query if none are suitable. 
We define each item's source as its URL's registrable domain and measure source preference by comparing source exposure and selection in the trajectories
(Implementation details are in \S\ref{app:agent-setting}).

\paragraph{Evaluation Domains.}
We evaluate agents across three domains where agents commonly make selections: Shopping, Accommodation, and Scholar.
To obtain realistic requests with specific requirements, we use WebShop \citep{yao2022webshop} for shopping products, HotelQuEST \citep{hadad2026hotelquest} for accommodation searches, and ScholarGym \citep{shen2026scholargym} for searching scholarly works.
We use 4,822 requests: 1,500 from WebShop, 786 from HotelQuEST, and 2,536 from ScholarGym. For Accommodation, we augment HotelQuEST's 214 requests with 572 location-substituted variants that preserve request structure (\S\ref{app:data-setting}).
We use these benchmarks only for requests, while item sets are constructed through the web search as described above. By evaluating user requests across diverse domains, we examine source preference in varied agent settings.

\paragraph{Requirement Satisfaction.}
We assess requirement satisfaction for every search result exposed to the agent using a source-blind judge (\S\ref{sec:measurement}; \S\ref{sec:relevance-utility}).
For Shopping, we use WebShop's structured requirements; for Accommodation and Scholar, an LLM extracts verifiable requirements from each request.
An LLM judge evaluates each requirement using only the result's title and content snippet, with URLs and source names removed from both.
For validation, human annotators provide requirement-level labels for a sample of agent-selected results in each domain.
Across domains, judge--human agreement is comparable to inter-annotator agreement (Krippendorff's $\alpha$: 0.65--0.75 vs.\ 0.67--0.72).
\S\ref{app:judge} provides methodological details and per-domain validation results.

\begin{figure*}[t]
    \centering
    \includegraphics[width=\textwidth]{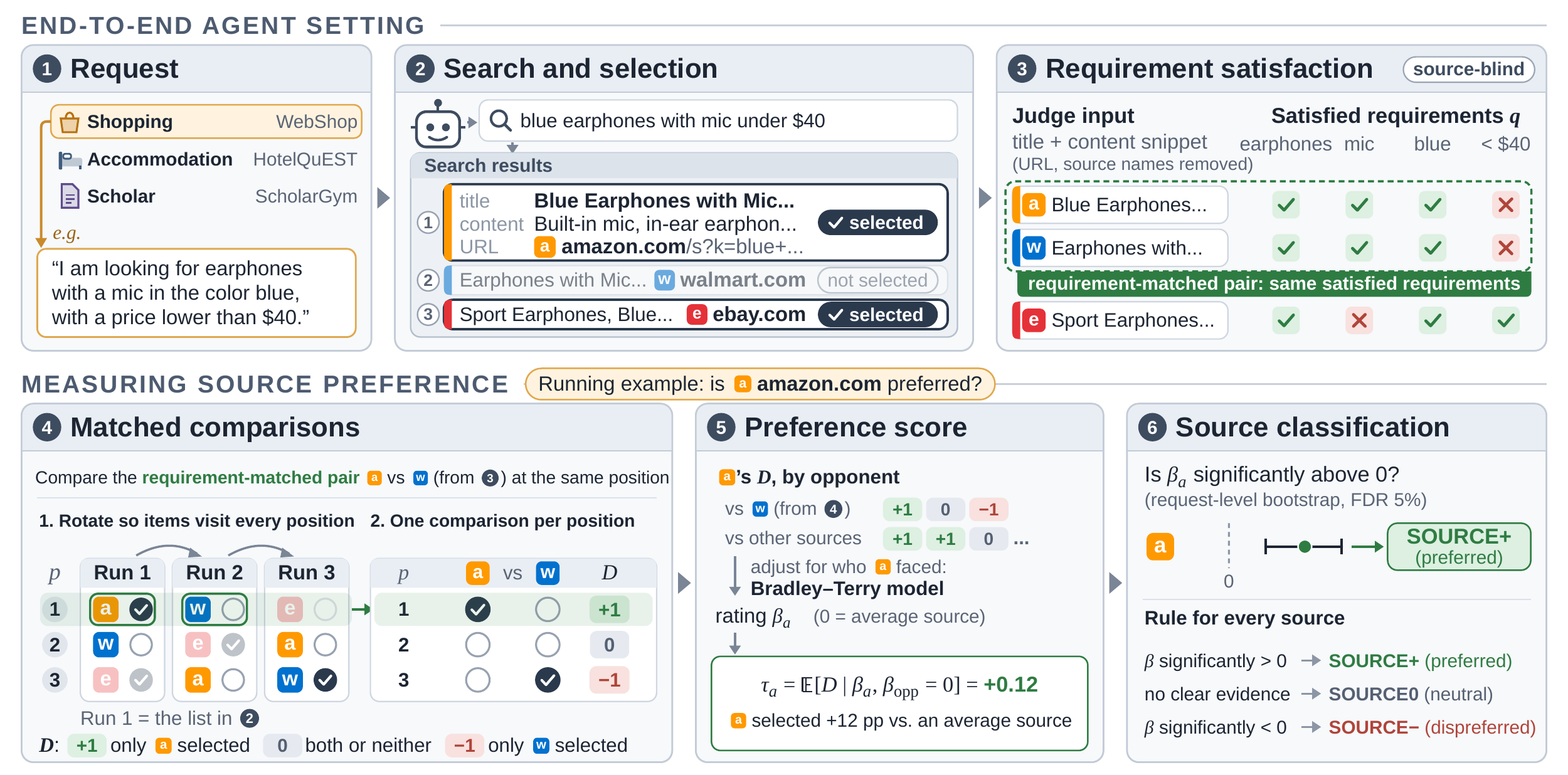}
\caption{\textbf{Overview.} Top: agent setting (\S\ref{sec:setting}). Bottom: measuring source preference (\S\ref{sec:measurement}).}
    \label{fig:source-preference-overview}
\end{figure*}

\section{Measuring Source Preference}
\label{sec:measurement}

To measure source preference, we compare how often items from different sources are selected under comparable conditions.
Specifically, we pair items that satisfy the same requirements and compare their selection outcomes at the same display position.
However, each source is compared with a different mix of other sources, making selection rates difficult to compare directly across sources.
To account for these differences, we fit a Bradley--Terry (BT) model~\citep{bradley1952rank} that estimates a rating for each source.
We convert each rating into a preference score, which summarizes how strongly the agent prefers the source.
Based on the significance of its rating, we then classify each source as preferred (\srcpos{}), dispreferred (\srcneg{}), or neutral (\srczero{}).

\paragraph{Matched Comparisons.}
We first pair items from different sources that satisfy the same requirements.
Let $q$ denote the set of requirements of the request that an item is judged to satisfy (\S\ref{sec:setting}).
Within the search result list for a request, two items from different sources with identical $q$ form a \emph{requirement-matched pair}, following matching designs in observational studies \citep{rubin1973matching,stuart2010matching}.
For example, in Fig.~\ref{fig:source-preference-overview}, panel~3, the Amazon and Walmart items form a requirement-matched pair because they have the same $q$, while the eBay item has a different $q$ and is not matched with either.
To compare the two items at the same display position, we present each list of length $L$ in all $L$ cyclic rotations, with each rotation evaluated in a separate run \citep{blankenstein2026biasbustersuncoveringmitigatingtool} (Fig.~\ref{fig:source-preference-overview}, panel~4).
Across these rotations, every item appears exactly once at each display position, forming a cyclic Latin square \citep{bailey2008design}.
We then compare the two items $i$ and $j$ of each requirement-matched pair position by position. For each position $p$, we record whether item $i$ is selected in the run that places $i$ at $p$, and whether item $j$ is selected in the run that places $j$ at $p$. These two outcomes form a \emph{comparison} (counts in \S\ref{app:bt}).

We index comparisons by $c$ and let $s_c$ and $s'_c$ denote the sources of items $i$ and $j$.
Let $Y_c, Y'_c\in\{0,1\}$ indicate whether $i$ and $j$, respectively, are selected in their respective runs.
We define the selection difference as
\(
D_c = Y_c - Y'_c \in \{-1,0,1\}.
\label{eq:comparison-outcome}
\)
$D_c=1$ means that $i$ is selected and $j$ is not; for $D_c=-1$, vice versa.
$D_c=0$ means that both items are selected or neither is selected.

\paragraph{Preference Score.}
With one source fixed as $i$, averaging $D_c$ over its comparisons against another source gives the difference in their selection rates.
To obtain one score per source $s$, we could average these differences across its comparisons, but the result would depend on which other sources it was compared with.
For example, a source may have a high average simply because it is mostly compared with sources the agent rarely selects.
We therefore use the BT model with Davidson's extension to account for ties ($D_c=0$) \citep{bradley1952rank,davidson1970extending,chiang2024chatbot}.
The model jointly estimates a rating $\beta_s$ for each source, accounting for which other sources it was compared with.
For comparison $c$, the probabilities of $D_c=1$, $-1$, and $0$ are proportional to $\exp(\beta_{s_c})$, $\exp(\beta_{s'_c})$, and $\nu\exp\!\left(\frac{\beta_{s_c}+\beta_{s'_c}}{2}\right)$, respectively, where $\nu>0$ controls the probability of a tie.
We estimate $\{\beta_s\}$ and $\nu$ by maximum likelihood with an $\ell_2$ penalty on the ratings for numerical stability. Sources with too few requirement-matched pairs to rate reliably are pooled into a single opponent with one shared rating, so their comparisons are still used (details and a fit check in \S\ref{app:bt}).
Since only differences between ratings matter, we center them so that the average rated source has $\beta=0$.
We then define each source's preference score as its expected selection difference $D$ against this average source:
\begin{equation}
\small
{
\tau_s
=
\mathbb{E}
\left[
D \mid \beta_s,\beta_{\mathrm{opp}}
\right]
=
\frac{\exp(\beta_s)-\exp(\beta_{\mathrm{opp}})}{\exp(\beta_s)+\exp(\beta_{\mathrm{opp}})+\nu\exp((\beta_s+\beta_\mathrm{opp})/2)}
=
\frac{\exp(\beta_s)-1}
{\exp(\beta_s)+1+\nu\exp(\beta_s/2)}}
\label{eq:tau-s}
\end{equation}
For example, $\tau_s=0.10$ means that the BT model predicts items from source $s$ to be selected 10 percentage points more often than those from the reference source (Fig.~\ref{fig:source-preference-overview}, panel~5).

\paragraph{Source Classification.}
We classify a source as preferred (\srcpos{}) when its rating is significantly positive, dispreferred (\srcneg{}) when it is significantly negative, and neutral (\srczero{}) otherwise (Fig.~\ref{fig:source-preference-overview}, panel~6).
These labels reflect statistical significance, not magnitude: \srczero{} means insufficient evidence of a preference in either direction, not a preference of zero.
To assess significance while accounting for dependence among comparisons from the same request, we use a request-level cluster bootstrap \citep{cameron2008bootstrap}.
In each replicate, we resample requests and refit the model.
From the bootstrap estimates, we obtain percentile confidence intervals for the source ratings and two-sided $p$-values for the null hypothesis $\beta_s=0$.
We control the false discovery rate at 5\% within each model and domain \citep{benjamini1995controlling} (robustness checks in \S\ref{app:bt}).

\section{Source Preference in the Wild}
\label{sec:natural}

\begin{table}[t]
\centering\small
\definecolor{pos}{HTML}{E58484}
\definecolor{neg}{HTML}{6D9EEB}
\newcommand{\pc}[2]{\cellcolor{pos!#1!white}$+#2$}
\newcommand{\nc}[2]{\cellcolor{neg!#1!white}$-#2$}
\newcommand{\zc}[1]{\textcolor{gray}{$#1$}}
\newcommand{\tbd}{\textcolor{gray}{tbd}}
\caption{Preference score $\tau_s$ (percentage points), with sources ordered by frequency; the row below the source names gives each source's share of all results, pooled over models. Cells are colored by label, determined after FDR correction (\S\ref{sec:measurement}): \srcpos{} \textcolor{pos}{red}, \srcneg{} \textcolor{neg}{blue}, with intensity proportional to $|\tau_s|$; \srczero{} entries are shown in \textcolor{gray}{gray}.}
% \vspace{\baselineskip}
\label{tab:wild-top}
\setlength{\tabcolsep}{3pt}
\renewcommand{\arraystretch}{1.15}
\resizebox{\linewidth}{!}{
\begin{tabular}{@{}llcccc@{\hspace{7pt}}cccc@{\hspace{7pt}}cccc@{}}
\toprule
& & \multicolumn{4}{c}{Shopping}
  & \multicolumn{4}{c}{Accommodation}
  & \multicolumn{4}{c}{Scholar} \\
\cmidrule(lr){3-6}\cmidrule(lr){7-10}\cmidrule(lr){11-14}
Family & Model
& Amazon & Walmart & eBay & Target
& Tripadvisor & Facebook & Expedia & Booking.com
& arXiv & ScienceDirect & OpenReview & Medium \\
& & {\scriptsize 12\%} & {\scriptsize 6\%}
  & {\scriptsize 5\%} & {\scriptsize 3\%}
  & {\scriptsize 9\%} & {\scriptsize 7\%}
  & {\scriptsize 4\%} & {\scriptsize 3\%}
  & {\scriptsize 12\%} & {\scriptsize 4\%}
  & {\scriptsize 4\%} & {\scriptsize 4\%} \\
\midrule
\textit{Qwen} & 3.5-27B
& \pc{52}{26} & \pc{40}{20} & \zc{+1} & \zc{+10}
& \zc{+2} & \nc{92}{46} & \nc{36}{18} & \pc{43}{21}
& \pc{65}{32} & \pc{28}{14} & \pc{36}{18} & \nc{48}{24} \\
& 3-30B-A3B
& \pc{31}{16} & \pc{42}{21} & \zc{+3} & \zc{+8}
& \nc{12}{6} & \nc{70}{35} & \nc{36}{18} & \pc{17}{8}
& \pc{47}{23} & \pc{15}{7} & \zc{+3} & \nc{26}{13} \\
& 2.5-32B
& \pc{31}{15} & \pc{33}{17} & \pc{15}{8} & \zc{+4}
& \nc{19}{10} & \nc{65}{32} & \nc{16}{8} & \pc{17}{8}
& \pc{46}{23} & \pc{16}{8} & \pc{15}{8} & \nc{13}{7} \\
\textit{GPT} & 5.4-nano
& \nc{12}{6} & \pc{22}{11} & \nc{36}{18} & \zc{+5}
& \nc{28}{14} & \nc{100}{52} & \zc{-5} & \pc{22}{11}
& \pc{67}{34} & \pc{22}{11} & \pc{57}{29} & \nc{47}{24} \\
& 5.6-Luna
& \pc{33}{16} & \pc{48}{24} & \pc{45}{23} & \pc{17}{8}
& \nc{24}{12} & \nc{83}{42} & \zc{-9} & \pc{31}{15}
& \pc{57}{28} & \pc{36}{18} & \pc{41}{21} & \nc{45}{23} \\
\textit{Llama} & 4-Maverick
& \pc{53}{27} & \pc{50}{25} & \zc{-1} & \zc{+7}
& \nc{22}{11} & \nc{86}{43} & \nc{43}{21} & \pc{19}{10}
& \pc{51}{25} & \zc{+2} & \zc{+1} & \nc{22}{11} \\
& 4-Scout
& \pc{37}{19} & \pc{41}{21} & \zc{+3} & \zc{+3}
& \nc{25}{13} & \nc{61}{31} & \nc{35}{18} & \zc{+7}
& \pc{56}{28} & \zc{-1} & \zc{+3} & \nc{22}{11} \\
& 3.1-8B-It
& \pc{12}{6} & \pc{23}{12} & \zc{+2} & \zc{+6}
& \nc{16}{8} & \nc{38}{19} & \zc{-3} & \pc{14}{7}
& \pc{31}{16} & \pc{6}{3} & \pc{11}{5} & \nc{10}{5} \\
& Tulu-3-8B
& \pc{10}{5} & \pc{22}{11} & \zc{+0} & \pc{24}{12}
& \nc{16}{8} & \nc{57}{28} & \nc{18}{9} & \pc{17}{9}
& \pc{39}{19} & \zc{+1} & \pc{6}{3} & \nc{11}{5} \\
\textit{Gemini} & 3.7-Flash
& \pc{12}{6} & \pc{22}{11} & \pc{23}{12} & \zc{-0}
& \zc{-4} & \nc{87}{43} & \zc{-6} & \zc{+6}
& \pc{58}{29} & \pc{29}{15} & \pc{43}{21} & \nc{52}{26} \\
\textit{GLM} & 5.3-Flash
& \pc{30}{15} & \pc{34}{17} & \pc{31}{16} & \zc{+8}
& \zc{+5} & \nc{100}{51} & \zc{-7} & \pc{31}{15}
& \pc{65}{33} & \pc{34}{17} & \pc{42}{21} & \nc{42}{21} \\
\textit{DeepSeek} & v4-Flash
& \pc{29}{14} & \pc{29}{15} & \pc{17}{9} & \zc{+5}
& \zc{-3} & \nc{98}{49} & \zc{-8} & \pc{27}{13}
& \pc{53}{27} & \pc{26}{13} & \pc{29}{14} & \nc{41}{21} \\
\bottomrule
\end{tabular}
}
\end{table}

This section applies the measure of \S\ref{sec:measurement} to end-to-end search. \S\ref{sec:which} reports which sources are preferred or dispreferred, across models and domains; \S\ref{sec:relevance-utility} asks how often preferred sources beat better items.

\subsection{Which Sources Are Preferred}
\label{sec:which}
All 12 models prefer some sources and avoid others in each of the three domains (Appendix Table~\ref{tab:cells}). 
Such preferences are also common among sources that appear most frequently in search results. 
Table~\ref{tab:wild-top} reports preference scores for the four most frequent sources in each domain, which together account for about a quarter of that domain's search results.
Of the 144 model–source combinations, 110 are classified as \srcpos{} or \srcneg{}, with median preference scores of $+15$ and $-18$ percentage points, respectively.
These preferences are widespread, but how consistent are they across models, and how do they vary across sources? 
First, models largely agree on which sources they prefer and avoid: for 10 of the 12 sources, no model prefers a source that another avoids.
The exceptions are Amazon and eBay, which GPT-5.4-nano avoids but other models prefer (more in \S\ref{app:wild}).
Second, preferences also differ among sources of the same type: all 12 models prefer Walmart, but only two prefer Target, another general retailer; ten prefer Booking.com, whereas half avoid Expedia, another hotel booking website.
Third, within the Qwen and Llama families, newer models exhibit stronger source preferences, as measured by mean $|\tau_s|$ across the 12 sources in Table~\ref{tab:wild-top}.

\begin{wrapfigure}{r}{0.5\linewidth}
\vspace{-\baselineskip}\centering
\includegraphics[width=0.8\linewidth]{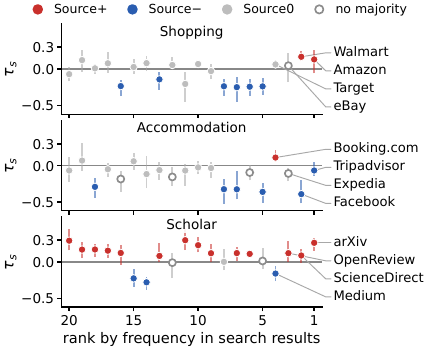}
\caption{Mean $\tau_s$ across models (points), with range (bars) and majority label (color).}
\label{fig:wild}
\vspace{-\baselineskip}
\end{wrapfigure}

Fig.~\ref{fig:wild} extends Table~\ref{tab:wild-top} to the twenty most frequent sources per domain, each colored by its majority label, the label assigned by more than half of the models that label it.
Agreement across models persists in this broader set: for 27 of the 32 red or blue sources, $\tau_s$ has the same sign for every model.
Among the 32, \srcpos{} dominates in Scholar (14 of 17), whereas \srcneg{} dominates in Shopping (6 of 8) and Accommodation (6 of 7).
Together, Table~\ref{tab:wild-top} and Fig.~\ref{fig:wild} show that models largely share which particular sources they prefer and avoid, even among sites of the same type, with the shared direction leaning toward \srcpos{} in Scholar and \srcneg{} in Shopping and Accommodation.

\subsection{How Often Preferred Sources Beat Better Items}
\label{sec:relevance-utility}

Items in a result list often differ in how well they satisfy the request. When the agent selects the less satisfying of two such items but not the more satisfying one, we call this an \textit{inversion}. We ask whether inversions are more common when the less satisfying item comes from a preferred source than from a dispreferred one.

Using the same runs, we form comparisons as in \S\ref{sec:measurement}, with one change: instead of requirement-matched pairs, we pair items from different sources in the same request that satisfy the same requirements except for one. We call the item that satisfies the extra requirement the \textit{more} satisfying item, and the other the \textit{less} satisfying item. We count only comparisons in which exactly one of the two items is selected. We call a pair an $(\ell, m)$ pair if the less satisfying item's source has label $\ell$ and the more satisfying item's source has label $m$. Then
\begin{equation}
\small
\mathrm{Inv}_{\ell,m}=\frac{\#\,\text{inversions}}{\#\,\text{comparisons with exactly one item selected}}\quad\text{among }(\ell,m)\text{ pairs}
\label{eq:inv}
\end{equation}
where $\ell, m \in \{\text{\srcpos}, \text{\srczero}, \text{\srcneg}\}$, written $+$, $0$, $-$ in subscripts. Thus $\mathrm{Inv}_{+,-}$ is the rate at which the less satisfying item from a preferred source is selected and the more satisfying item from a dispreferred source is not; $\mathrm{Inv}_{-,+}$ is the corresponding rate with the source labels reversed.
Without source preference, the source would not affect which item is selected, so $\mathrm{Inv}_{+,-}$ and $\mathrm{Inv}_{-,+}$ would be similar; the gap between them thus shows how far source preference overrides the difference in satisfied requirements. $\mathrm{Inv}_{0,0}$, for pairs of neutral sources, serves as a baseline.
\begin{figure}[t]\centering
\includegraphics[width=0.9\linewidth]{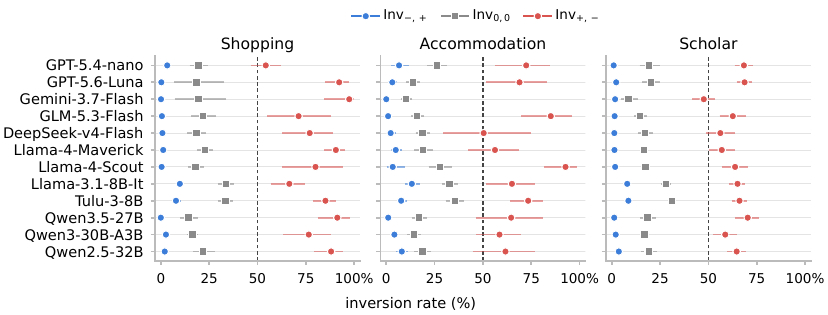}
\caption{Inversion rates of Eq.~\ref{eq:inv}. Bars: 95\% bootstrap confidence intervals; dashed line: 50\%. Each point rests on a median of about 2{,}000 comparisons with exactly one item selected; the missing point has no pairs to compare. Exact values in Table~\ref{tab:inversion-n}.}
\label{fig:inversion}
\end{figure}

Fig.~\ref{fig:inversion} shows a sharp contrast depending on which item comes from the preferred source. The less satisfying item is selected 68\% of the time at the median when it comes from a preferred source and the other from a dispreferred source ($\mathrm{Inv}_{+,-}$);
when reversed ($\mathrm{Inv}_{-,+}$), it is almost never selected (median 2\%). Neutral-source pairs ($\mathrm{Inv}_{0,0}$, median 19\%) fall in between, and this order holds across model--domain combinations.
Separating the two sides by pairing each with a neutral source shows that the dispreferred side matters more: in every domain, most models show more inversions with a dispreferred source ($\mathrm{Inv}_{0,-}$) than with a preferred one ($\mathrm{Inv}_{+,0}$); per-model values are in \S\ref{app:wild}.
Overall, inversions are more common when the more satisfying item comes from a dispreferred source than when the less satisfying item comes from a preferred source.

\section{Source Preference in Controlled Experiments}
\label{sec:origin}

In \S\ref{sec:natural}, agents show source preference even between items matched on the requirements they satisfy.
However, a preference for items from certain sources does not by itself show that agents' choices depend on explicit source information---URLs and source names in the title and content.
The preference may instead reflect other differences between the items.
We therefore test, in two controlled experiments, whether explicit source information itself affects selection.
If it does, hiding it should weaken the preference, and changing it should change which item is selected.
We first examine how source preference changes when this information is hidden and then restored in the same result lists, while keeping all other information unchanged (\S\ref{sec:visibility}).
We then test whether the same items are selected differently when we swap their explicit source information between preferred and dispreferred sources, while keeping their title and content fixed (\S\ref{sec:controlled-source-swap}).

\subsection{Removing Explicit Source Information Weakens Preference}
\begin{wraptable}{r}{0.50\textwidth}
\vspace{-1\baselineskip}
\centering
\small
\caption{\textbf{Source-disclosure conditions.} All con-
ditions use the same search results}
\label{tab:conditions}
\resizebox{0.8\linewidth}{!}{%
\begin{tabular}{@{}lll@{}}
\toprule
Condition & Title / content & URL \\
\midrule
\chid{}  & source names masked & hidden \\
\cred{}  & source names masked & shown \\
\corig{} & original text            & shown \\
\bottomrule
\end{tabular}
}
\vspace{-0.8em}
\end{wraptable}
\label{sec:visibility}

To assess how revealing explicit source information changes source preference, we compare agents' selections from the same search result lists with explicit source information hidden, then restored in two steps.
This comparison measures how much preference remains when explicit source information is hidden and how preference changes as URLs and then source names in the text are restored.
In Table~\ref{tab:conditions}, \chid{} hides explicit source information in both the URL and text, \cred{} exposes the URL while masking source names in the text, and \corig{} uses the unmodified search results from \S\ref{sec:natural}.
We estimate the preference score $\tau_s$ separately under each condition: the change from \chid{} to \cred{} measures the effect of revealing the URL, while the change from \cred{} to \corig{} measures the additional effect.
We keep the source classifications from \S\ref{sec:natural} fixed across conditions and, for each model and source group, report the macro-average of the three domain-level mean scores.
See details in \S\ref{app:source-visibility-details}.

\begin{figure*}[t]
\centering
\small
\includegraphics[width=0.9\textwidth]{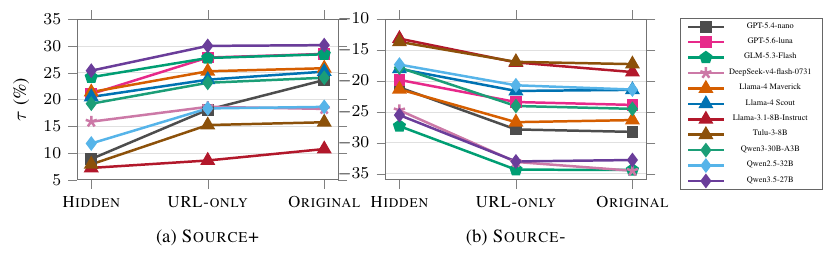}
\caption{\textbf{Source preference across disclosure conditions.}
Each line shows a model's preference score $\tau$, averaged equally across the three domains and expressed in percentage points, for (a) \srcpos{} or (b) \srcneg{}.
The \chid{}$\rightarrow$\cred{} transition restores the URL; the \cred{}$\rightarrow$\corig{} transition additionally restores source names in the title and content.}
\label{fig:oof-source-preference-decomposition}
\end{figure*}

Fig.~\ref{fig:oof-source-preference-decomposition} shows that restoring explicit source information (\chid{}$\rightarrow$\corig{}) increases $\tau$ for \srcpos{} and decreases it for \srcneg{} in all models. 
These changes are approximately symmetric on average: $\tau$ rises by 5.8 percentage points for \srcpos{} and falls by 5.9 points for \srcneg{}, averaging equally across models.
Thus, revealing URLs and source names widens the gap in preference scores between \srcpos{} and \srcneg{} compared with hiding them.
Most of this widening occurs when the URL alone is revealed, with marginal change when source names in the text are restored.
In nine of eleven models, over 80\% of the total gap widening from \chid{} to \corig{} occurs in the first step, when only the URL is restored (\chid{}$\rightarrow$\cred{}).

\subsection{Changing Explicit Source Information Changes Selection}
\label{sec:controlled-source-swap}

\begin{wraptable}{r}{0.5\textwidth}
\vspace{-1.6\baselineskip}  
\centering
\small
\definecolor{swappos}{HTML}{E58484}
\setlength{\abovecaptionskip}{0pt}
\setlength{\belowcaptionskip}{6pt}
\caption{\textbf{Effect of displayed explicit source information.}
Values show \(T_+ - T_-\) (pp): the increase in
selection under \srcpos{} versus \srcneg{}
when the two sources are swapped.
\(\dagger\): \(p<0.01\); \(\ddagger\): \(p<0.001\) against zero.}
\vspace{\baselineskip}
\label{tab:source-swap-tie}
\setlength{\tabcolsep}{4pt}\renewcommand{\arraystretch}{1.20}
\resizebox{0.8\linewidth}{!}{%
\begin{tabular}{@{}lccc@{}}
\toprule
Model & Shopping & Accommodation & Scholar \\
\midrule
GPT-5.4-nano & \cellcolor{swappos!19.17!white}\(+19.2^{\ddagger}\) & \cellcolor{swappos!24.05!white}\(+24.1^{\ddagger}\) & \cellcolor{swappos!18.63!white}\(+18.6^{\ddagger}\) \\
GPT-5.6-Luna & \cellcolor{swappos!26.95!white}\(+27.0^{\ddagger}\) & \cellcolor{swappos!20.13!white}\(+20.1^{\ddagger}\) & \cellcolor{swappos!14.07!white}\(+14.1^{\ddagger}\) \\
GLM-5.3-Flash & \cellcolor{swappos!25.81!white}\(+25.8^{\ddagger}\) & \cellcolor{swappos!12.74!white}\(+12.7^{\ddagger}\) & \cellcolor{swappos!10.40!white}\(+10.4^{\ddagger}\) \\
DeepSeek-v4-Flash & \cellcolor{swappos!26.46!white}\(+26.5^{\ddagger}\) & \cellcolor{swappos!11.82!white}\(+11.8^{\ddagger}\) & \cellcolor{swappos!9.66!white}\(+9.7^{\ddagger}\) \\
Gemini-3.7-Flash & \cellcolor{swappos!38.51!white}\(+38.5^{\ddagger}\) & \cellcolor{swappos!13.33!white}\(+13.3\) & \cellcolor{swappos!13.53!white}\(+13.5^{\ddagger}\) \\
Llama-4-Maverick & \cellcolor{swappos!64.71!white}\(+64.7^{\ddagger}\) & \cellcolor{swappos!14.92!white}\(+14.9^{\ddagger}\) & \cellcolor{swappos!21.40!white}\(+21.4^{\ddagger}\) \\
Llama-4-Scout & \cellcolor{swappos!48.83!white}\(+48.8^{\ddagger}\) & \cellcolor{swappos!2.46!white}\(+2.5\) & \cellcolor{swappos!16.12!white}\(+16.1^{\ddagger}\) \\
Llama-3.1-8B-It & \cellcolor{swappos!43.73!white}\(+43.7^{\ddagger}\) & \cellcolor{swappos!4.09!white}\(+4.1\) & \cellcolor{swappos!11.84!white}\(+11.8^{\ddagger}\) \\
Tulu-3-8B & \cellcolor{swappos!58.71!white}\(+58.7^{\ddagger}\) & \cellcolor{swappos!11.24!white}\(+11.2^{\ddagger}\) & \cellcolor{swappos!10.53!white}\(+10.5^{\ddagger}\) \\
Qwen3.5-27B & \cellcolor{swappos!43.71!white}\(+43.7^{\ddagger}\) & \cellcolor{swappos!8.56!white}\(+8.6^{\ddagger}\) & \cellcolor{swappos!14.99!white}\(+15.0^{\ddagger}\) \\
Qwen3-30B-A3B & \cellcolor{swappos!41.43!white}\(+41.4^{\ddagger}\) & \cellcolor{swappos!6.25!white}\(+6.3^{\dagger}\) & \cellcolor{swappos!15.45!white}\(+15.4^{\ddagger}\) \\
Qwen2.5-32B & \cellcolor{swappos!32.46!white}\(+32.5^{\ddagger}\) & \cellcolor{swappos!5.90!white}\(+5.9^{\ddagger}\) & \cellcolor{swappos!8.83!white}\(+8.8^{\ddagger}\) \\
\bottomrule
\end{tabular}
}
\end{wraptable}

In \S\ref{sec:visibility}, we show that revealing URLs and source names strengthens source preference.
However, this comparison does not directly establish whether the same item would be selected differently under different displayed sources.
We therefore ask whether the same item is more likely to be selected when its displayed source is \srcpos{} rather than \srcneg{}.
To test this, we select requirement-matched pairs, one item from \srcpos{} and one from \srcneg{}.
For each pair, we keep both items' title and content fixed and compare two conditions: one with the original sources and the sources swapped.
We present the pair in both orders under each condition, giving four evaluations per pair~\citep{senn2002crossover}.
Each item therefore appears with both \srcpos{} and \srcneg{}, once in each position.
Let $T_+$ and $T_-$ be the proportions of choices for items displaying \srcpos{} and \srcneg{}, respectively.
The difference, $T_+ - T_-$, measures how much more often an item is selected when it displays \srcpos{} than \srcneg{}.
Since content and position are balanced across sources, preferences for either alone would give an expected difference of zero.
A positive difference means displaying \srcpos{} increases the likelihood of selection.
Detailed settings are provided in \S\ref{app:source-swap-details}.

Table~\ref{tab:source-swap-tie} shows that the same content is selected more often when it displays \srcpos{} rather than \srcneg{}.
The difference $T_+ - T_-$ is positive across models and domains, and statistically significant at \(p<0.01\) in at least one domain for all 12 models.
Whereas \S\ref{sec:visibility} shows that revealing source information strengthens source preference, this experiment shows that changing the explicit source information changes selection when content is held fixed.
Beyond the difference in selection rates depending on whether the preferred source is shown or hidden (\S\ref{sec:visibility}), the same content is selected more often when its source is changed from \srcneg{} to \srcpos{}.
With content and position balanced across source assignments, these results provide evidence that displaying \srcpos{} rather than \srcneg{} increases selection in this controlled setting.

\section{Source Preference Shaped by Training Data}
\label{sec:dpo}
% ---------- Document ----------
% Place after the subsection heading, immediately before its body text.
\begin{wraptable}{r}{0.5\textwidth}
\centering
\captionsetup{font=small,position=bottom,skip=6pt}
\caption{\textbf{Target-source selection rate (\%) on
equal-satisfaction test pairs, averaged across three fake source pairs.}
Entries report the mean $\pm$ sample standard deviation across fake
source pairs. Balanced, Aligned, and Reversed pair the target source
with the preferred response in 50\%, 80\%, and 20\% of training pairs,
respectively.}
\vspace{\baselineskip}
% ---------- Table 4: fake sources ----------
{\scriptsize
\setlength{\tabcolsep}{3pt}
\resizebox{\linewidth}{!}{%
  \begin{tabular}{@{}lcccc@{}}
    \toprule
    & Pre-DPO & \multicolumn{3}{c}{DPO ($n=5{,}000$)} \\
    \cmidrule(l){3-5}
    Model & ($n=0$) & Balanced & Aligned & Reversed \\
    \midrule
    Qwen3.5-9B
    & $50.6 \pm 1.0$ & $50.1 \pm 0.5$
    & $70.1 \pm 0.5$ & $28.5 \pm 1.4$ \\
    Llama-3.1-8B
    & $49.8 \pm 0.4$ & $49.4 \pm 0.1$
    & $73.8 \pm 0.4$ & $26.5 \pm 0.5$ \\
    Tulu-3-8B-SFT
    & $51.0 \pm 2.3$ & $50.1 \pm 0.4$
    & $75.1 \pm 0.6$ & $24.5 \pm 1.0$ \\
    \bottomrule
  \end{tabular}%
}
\par}

\label{tab:kelto-choice-rate}

\par\vspace{10pt}
\caption{\textbf{Amazon selection rate (\%) on
equal-satisfaction test pairs.}
Balanced, Aligned, and Reversed pair Amazon with the preferred
training response in 50\%, 80\%, and 20\% of pairs, respectively.}
\vspace{\baselineskip}
% ---------- Table 5: real sources ----------
{\scriptsize
\setlength{\tabcolsep}{4pt}
\resizebox{\linewidth}{!}{%
  \begin{tabular}{@{}lrrrr@{}}
    \toprule
    & Pre-DPO & \multicolumn{3}{c}{DPO ($n=5{,}000$)} \\
    \cmidrule(l){3-5}
    Model & ($n=0$) & Balanced & Aligned & Reversed \\
    \midrule
    \multicolumn{5}{@{}l}{\textit{Qwen3.5-9B}} \\
    \quad vs.\ eBay       & 54.1 & 50.0 & 79.3 & 24.5 \\
    \quad vs.\ Etsy       & 60.5 & 51.8 & 79.0 & 23.8 \\
    \quad vs.\ AliExpress & 60.0 & 49.7 & 80.6 & 25.3 \\
    \addlinespace
    \multicolumn{5}{@{}l}{\textit{Llama-3.1-8B}} \\
    \quad vs.\ eBay       & 53.7 & 50.2 & 78.3 & 24.8 \\
    \quad vs.\ Etsy       & 56.8 & 51.9 & 79.2 & 23.3 \\
    \quad vs.\ AliExpress & 54.6 & 51.1 & 78.0 & 24.5 \\
    \addlinespace
    \multicolumn{5}{@{}l}{\textit{Tulu-3-8B-SFT}} \\
    \quad vs.\ eBay       & 58.0 & 51.1 & 82.2 & 19.3 \\
    \quad vs.\ Etsy       & 62.7 & 51.0 & 82.5 & 17.7 \\
    \quad vs.\ AliExpress & 59.5 & 51.0 & 82.4 & 18.4 \\
    \bottomrule
  \end{tabular}%
}
\par}

\label{tab:amazon-choice-rate}

\end{wraptable}
\subsection{Creating Source Preference from Source–Satisfaction Correlation}
\label{sec:training}
In our results, sources whose items satisfy user requirements well tend to receive higher preference scores (details in \S\ref{app:source-satisfaction-correlation}).
Models may have encountered a similar relationship during training.
If products from one source are repeatedly preferred because they better satisfy requirements, the agent may learn to associate that source with better items.
The agent may then rely on explicit source information rather than assess how well each product satisfies the user's requirements.
The agent may thus learn to use the source as a shortcut~\citep{geirhos2020shortcut}, favoring it even when products from different sources satisfy the request equally well.
We test whether this training-time association creates a source preference that persists even when the compared items satisfy the same requirements.

% \begin{wraptable}{r}{0.52\textwidth}
% \vspace{0pt}  
% \centering
% \scriptsize
% \caption{\textbf{Target-source selection rate (\%) on equal-satisfaction test pairs, averaged across three fake source pairs.}
% Entries report the mean $\pm$ sample standard deviation across fake source pairs. Balanced, Aligned, and Reversed pair the target source with the preferred response in 50\%, 80\%, and 20\% of training pairs, respectively.}
% \vspace{\baselineskip}
% \label{tab:kelto-choice-rate}
% \setlength{\tabcolsep}{3pt}
% \resizebox{0.9\linewidth}{!}{%
% \begin{tabular}{@{}lcccc@{}}
% \toprule
%  & Pre-DPO & \multicolumn{3}{c}{DPO ($n=5{,}000$)} \\
% \cmidrule(l){3-5}
% Model & ($n=0$) & Balanced & Aligned & Reversed \\
% \midrule
% Qwen3.5-9B
%   & $50.6 \pm 1.0$ & $50.1 \pm 0.5$ & $70.1 \pm 0.5$ & $28.5 \pm 1.4$ \\
% Llama-3.1-8B
%   & $49.8 \pm 0.4$ & $49.4 \pm 0.1$ & $73.8 \pm 0.4$ & $26.5 \pm 0.5$ \\
% Tulu-3-8B-SFT
%   & $51.0 \pm 2.3$ & $50.1 \pm 0.4$ & $75.1 \pm 0.6$ & $24.5 \pm 1.0$ \\
% \bottomrule
% \end{tabular}
% }
% \vspace{-2em}
% \end{wraptable}

We train Qwen3.5-9B, Llama-3.1-8B-Instruct, and Tulu-3-8B-SFT to select one of two products using Direct Preference Optimization (DPO) \citep{rafailov2024directpreferenceoptimizationlanguage}.
We run the experiment using three pairs of unfamiliar, fake sources: \textit{Kelto--Varnis}, \textit{Temnuki--Dalnuki}, and \textit{Roveki--Tavumi}. 
In each pair, the first source is designated as the target.
In each DPO pair, the preferred response selects the product that better satisfies the request, and the other response selects the other product.
The target source labels the better product in 50\%, 80\%, and 20\% of pairs in \textit{Balanced}, \textit{Aligned}, and \textit{Reversed}, respectively.
Explicit source information thus has no association, a positive association, or a negative association with the preferred response during training.
At inference, each pair contains two equally satisfying products, one from each unfamiliar source, so explicit source information provides no information about which product better satisfies the request; i.e., this tests whether the agent favors a source even when it offers no advantage in satisfying the request.
All products are retrieved for WebShop requests, with their original source information removed before the fake sources are assigned.
Details are provided in \S\ref{app:dpo}.

% \begin{wraptable}[18]{r}{0.5\linewidth}
% \vspace{-2\baselineskip}  
% \centering
% \small
% \caption{\textbf{Amazon selection rate (\%) on equal-satisfaction test pairs.}
% Balanced, Aligned, and Reversed pair Amazon with the preferred training
% response in 50\%, 80\%, and 20\% of pairs, respectively.}
% \label{tab:amazon-choice-rate}
% \setlength{\tabcolsep}{4pt}
% \resizebox{0.9\linewidth}{!}{%
% \begin{tabular}{@{}lrrrr@{}}
% \toprule
% & Pre-DPO & \multicolumn{3}{c}{DPO ($n=5{,}000$)} \\
% \cmidrule(l){3-5}
% Model & ($n=0$) & Balanced & Aligned & Reversed \\
% \midrule
% \multicolumn{5}{@{}l}{\textit{Qwen3.5-9B}} \\
% \quad vs.\ eBay       & 54.1 & 50.0 & 79.3 & 24.5 \\
% \quad vs.\ Etsy       & 60.5 & 51.8 & 79.0 & 23.8 \\
% \quad vs.\ AliExpress & 60.0 & 49.7 & 80.6 & 25.3 \\
% \addlinespace
% \multicolumn{5}{@{}l}{\textit{Llama-3.1-8B}} \\
% \quad vs.\ eBay       & 53.7 & 50.2 & 78.3 & 24.8 \\
% \quad vs.\ Etsy       & 56.8 & 51.9 & 79.2 & 23.3 \\
% \quad vs.\ AliExpress & 54.6 & 51.1 & 78.0 & 24.5 \\
% \addlinespace
% \multicolumn{5}{@{}l}{\textit{Tulu-3-8B-SFT}} \\
% \quad vs.\ eBay       & 58.0 & 51.1 & 82.2 & 19.3 \\
% \quad vs.\ Etsy       & 62.7 & 51.0 & 82.5 & 17.7 \\
% \quad vs.\ AliExpress & 59.5 & 51.0 & 82.4 & 18.4 \\
% \bottomrule
% \end{tabular}%
% }
% \end{wraptable}

Table~\ref{tab:kelto-choice-rate} shows that models learn to favor or avoid a source depending on how often its products are preferred during training.
Averaged across three fake source pairs, the target-source selection rate remains near 50\% before DPO and under \textit{Balanced} training (49.4--51.0\%).
It rises to 70.1--75.1\% under \textit{Aligned} and falls to 24.5--28.5\% under \textit{Reversed}.
With equally satisfying products at inference, these shifts reflect source preference rather than differences in requirement satisfaction.
This implies that the training can create source preference by repeatedly pairing sources with the preferred response, although labels depend solely on requirement satisfaction.

\subsection{Mitigating an Existing Source Preference by Rebalancing Training Data}
\label{sec:training-mitigation}

The fake-source experiment (\S\ref{sec:training}) shows that preference training can create a source preference when a source repeatedly accompanies the preferred response.
We next test whether changing this association can weaken an existing preference for a familiar source.
We compare Amazon with eBay, Etsy, and AliExpress; before DPO, all three models favor Amazon over each of them on equal-satisfaction test pairs.
Table~\ref{tab:amazon-choice-rate} shows that preference training offers a possible way to reduce existing source preferences.
Before DPO, Amazon selection rates range from 53.7\% to 62.7\%.
Balanced DPO moves them close to 50\% (49.7--51.9\%), while Aligned DPO raises them to 78.0--82.5\%.
Reversed DPO lowers them to 17.7--25.3\%, favoring the previously less-preferred source for every compared source.
These results suggest that changing which source accompanies preferred responses can weaken or reverse an existing source preference.

\section{Source Preference under Missing Information}
\label{sec:rationale-rationale}

When an item's content lacks information the request asks for, agents tend to fill the gap from the source.
Specifically, when a request specifies a price that the item's content omits, we observe agents justifying the selection of a preferred source's item by reasoning that it usually offers low prices (details in \S\ref{app:agent-reasoning}).
We hypothesize that such missing information triggers the agent's preconceptions about the source, strengthening its source preference. To test this, we follow the two-item setting of \S\ref{sec:controlled-source-swap}: requirement-matched pairs of products retrieved for WebShop requests, one product from \srcpos{} and one from \srcneg{}, presented in both orders. We measure the selection rate of the \srcpos{} item for the five models in Table~\ref{tab:winrate}, and intervene in two ways: supplying the missing price (\S\ref{sec:information-suff}) and controlling what the agent assumes about the source (\S\ref{sec:practical-mitgation}).

\begin{wraptable}{r}{0.55\linewidth}
\vspace{-\baselineskip}
\centering
\small
\caption{\textbf{Source preference under added information or instructions.}
Base reports the \srcpos{} selection rate.
Each intervention is applied independently;
changes are measured in percentage points (pp).}
\label{tab:winrate}
\setlength{\tabcolsep}{4pt}
\resizebox{0.9\linewidth}{!}{%
\begin{tabular}{@{}lrrrr@{}}
\toprule
& Base (\%) & \multicolumn{3}{c}{Change from Base (pp)} \\
\cmidrule(l){3-5}
Model & & Same Price & General & Designate \\
\midrule
GPT-5.4-nano
& 68.8 & $-3.9$  & $-1.1$ & $-17.1$ \\
Llama-4-Maverick
& 95.7 & $-7.1$  & $+1.0$ & $-12.5$ \\
Llama-3.1-8B-It
& 83.9 & $-17.7$ & $-0.6$ & $-7.4$  \\
Qwen3.5-27B
& 96.4 & $-2.8$  & $+1.3$ & $-22.5$ \\
Tulu-3-8B
& 87.6 & $-28.3$ & $-0.1$ & $-11.8$ \\
\bottomrule
\end{tabular}%
}
\vspace{0pt}
\end{wraptable}

\subsection{Supplying the Missing Information}
\label{sec:information-suff}
To test whether supplying the missing information weakens source preference, we compare two settings: \textit{Base}, the setting above without intervention, and \textit{Same Price}, which adds to the content of both products an identical price that satisfies the requirement. Table~\ref{tab:winrate} shows that \textit{Same Price} lowers the selection rate of the \srcpos{} item in all models. That merely supplying the missing information weakens the preference suggests the agent indeed relies on its preconceptions about the source when information is missing.

\subsection{Mitigating Source Preference through the System Prompt}
\label{sec:practical-mitgation}
Supplying the missing information reduces the preference, but in practice it is missing because it is unavailable. We therefore control, through the system prompt, what the agent assumes about the source when information is missing.
We test two system-prompt instructions: \textit{General} states, ``The tagged source URL has no relation whatsoever to the product's price.'' \textit{Designate} states, ``Usually, \{retailer\} serves cheap price product,'' where \{retailer\} is replaced with the \srcneg{} retailer, countering the preconception in favor of \srcpos{}.
Table~\ref{tab:winrate} shows that \textit{General} leaves the selection rate of the \srcpos{} item nearly unchanged, whereas \textit{Designate} lowers it in every model. Removing the preconception has little effect, whereas replacing it reduces the preference. This counteracts the preference rather than removing its cause and requires knowing which source is dispreferred, but it shows that controlling what the agent assumes about the source can move the preference.

\section{Conclusion}
We studied source preference in end-to-end search across 12 agent models and three domains. 
Agents favor sources even when items satisfy the same requirements, and these preferences can accompany selections of less satisfying items. 
Controlled experiments establish that source identity itself affects selection, while preference training shows how source–satisfaction associations can create or reduce source preference. 
Supplying missing information and prompting agents to reconsider source-based assumptions reduce preference. 
These findings highlight the need to evaluate agents not only by which items they select, but also by how source identity shapes those selections.

\subsection*{AI use statement}
In this work, we used generative AI tools to propose and refine hypotheses, design and provide feedback on research methodology and experiments, implement methods, assist with translation, support qualitative and thematic data analysis, generate synthetic datasets, and interpret results. We have not used generative AI tools to help develop theoretical models or conceptual frameworks, or clean and reformat datasets. Formulating mathematical claims, providing critical ingredients for proving mathematical claims, and assisting in the writing of proofs are not applicable to this work. We have reviewed all AI-assisted work. All AI-suggested hypotheses and methodological choices were critically evaluated and finalized by the authors, all AI-generated code was reviewed and tested for correctness by the authors, and all AI-assisted translations were reviewed by the authors. All AI-assisted qualitative analyses and interpretations of results were verified by the authors against the underlying data. We take responsibility for the final content of this work, including text, claims, or artifacts produced with the aid of generative AI.

\subsection*{Reproducibility statement}
We describe the search-and-selection workflow and evaluation data in \S\ref{sec:setting}, and the matched comparisons, position controls, and source-preference measure in \S\ref{sec:measurement}. The appendices provide the search and agent configurations (\S\ref{app:agent-setting}), data construction procedure (\S\ref{app:data-setting}), and request-satisfaction judging procedure and human validation (\S\ref{app:judge}). For deterministic decoding, we set the temperature and seed for every model and checked whether the same response is returned from the open-weight models. We will also release the code and dataset upon publication.

\bibliography{iclr2027_conference}
\bibliographystyle{iclr2027_conference}
\newpage
\appendix

\appendix
\counterwithin{table}{section}
\counterwithin{figure}{section}

\section{Related Works}

\paragraph{Preference and Bias in LLMs}
Prior work has examined how answer position and contextual cues influence LLM choices.
Prior studies showed that changing the order of answer options systematically affects model performance \citep{pezeshkpour-hruschka-2024-large,zheng2024largelanguagemodelsrobust}.
Beyond position, cues such as information source, authority, and social proof can also shift model choices ~
\citep{mammen2026endorseditmeasuringauthority, filandrianos2025biasbewareimpactcognitive}.
In retrieval-augmented generation, prior work examined how models' preferences for information vary with its source \citep{li-etal-2025-llms-trust}.
These findings extend to agents that select items, information sources, and tools.
Prior work found source preferences across academic paper, news, and shopping domains \citep{khan2026agents}, while another study examined how price, ratings, and nudging messages influence shopping agents' choices \citep{cherep2026frameworkstudyingaiagent}. Even among functionally equivalent tools, agents exhibit systematic selection biases associated with the provider and the tool's position in the context \citep{blankenstein2026biasbustersuncoveringmitigatingtool}.
Our work extends the study of source preference to end-to-end web search, where agents form their own item sets through search.
We measure source preference among items with equal request satisfaction while controlling for position, and use source-name swaps with content held fixed to isolate the effect of displayed explicit source information.

% \paragraph{Provider Fairness and Exposure Allocation}
% Source preference also matters for how opportunities are distributed across providers.
% In recommender systems, this concern has motivated research on fairness toward both users and providers, including how recommendation systems distribute exposure across items and suppliers ~\citep{burke2017multisidedfairnessrecommendation,Patro_2020,Mansoury_2021}.
% As LLMs take on recommendation tasks, related concerns arise about how their choices distribute exposure. Prior work examined how prompt structure, system role, and intent affect provider fairness and catalog coverage in ChatGPT-based recommenders ~\citep{deldjoo2024understandingbiaseschatgptbasedrecommender}.
% For agent-based recommendation, a tri-party framework was proposed that coordinates user utility, item exposure, and platform fairness through item self-promotion and platform reranking ~\citep{gong2026breakingusercentricagencytriparty}.
% While these studies examine provider exposure and fairness in recommendation, our work analyzes whether source preference disadvantages providers whose candidates better satisfy user requests in end-to-end web search.
% We further examine how source preference in LLMs translates into provider disadvantage and how system prompts and preference training can mitigate it.

\paragraph{Provider Fairness and Exposure Allocation}
Fairness in recommender systems is usually framed as a multi-stakeholder problem. Besides serving users, a recommender decides how much exposure each item and supplier gets~\citep{burke2017multisidedfairnessrecommendation,Patro_2020,Mansoury_2021}. The same question now applies to agent recommender. \citet{deldjoo2024understandingbiaseschatgptbasedrecommender} show that prompt structure, system role, and user's intent affect provider fairness and catalog coverage in ChatGPT-recommender. In agent system, \citet{gong2026breakingusercentricagencytriparty} let items promote themselves and have the platform rerank them, balancing user utility, item exposure, and platform fairness. However, existing works quantify exposure within a recommendation list. In our work, the setting is end-to-end web search, where the question is whether an LLM's source preference passes over the provider whose item satisfies the user's request. Also, we measure how large this disadvantage is and test whether system prompts and preference training reduce it.

\section{End-to-End Web Agent Details}
\label{app:agent-setting}
\paragraph{Search Configuration.}
All agents use the Tavily Search API to retrieve web results. Each search sends one request with \texttt{topic=general}, \texttt{search\_depth=basic}, and \texttt{max\_results=10}. For each result, the agent is shown the title, the content snippet provided by Tavily, and the URL, in that order. We present the results in the order returned by Tavily, without re-ranking or shuffling, and number them \texttt{[1]}--\texttt{[k]}.
Tavily does not always return the requested maximum of 10 results, thereby the average item number per search result was 8.83 (WebShop: 8.59, Accommodation: 9.04, Scholar: 8.91).

\paragraph{Agent Actions and Interaction Limits.}
Our environment is text-only and has two pages: a \emph{search page} and a \emph{results page}. The agent can take the following actions:
\begin{itemize}
    \item \texttt{search[<query>]} (search page only): submits a free-form query to Tavily and shows the results page.
    \item \texttt{select[<n1,n2,...>]} (results page only): chooses any subset of the displayed results (e.g., \texttt{select[2]} or \texttt{select[1,4,7]}). This ends the episode immediately. There is no separate purchase or confirmation step.
    \item \texttt{select[Back to Search]} (results page only): returns to the search page so that the agent can issue a new query.
\end{itemize}
The agent cannot open, click into, or browse any webpage. The title, snippet, and URL on the results page are all the evidence it has.
An episode may contain at most five environment actions. Each additional search takes two actions (\texttt{Back to Search} followed by \texttt{search}), so an agent can make at most three searches per episode, and at most two if it still needs to make a selection (e.g., \texttt{search} $\rightarrow$ \texttt{Back} $\rightarrow$ \texttt{search} $\rightarrow$ \texttt{select}).
The agent can select only from the most recent results page.
Malformed or inadmissible actions (e.g., a missing \texttt{<action>} tag, or \texttt{search} on the results page) return a corrective error message and still count toward the budget.
An empty selection (\texttt{select[]}) is allowed and is treated as selecting none of the items on the search results page.

\paragraph{Agent Prompt and Model Configuration.}
\label{app:agent-prompt}
The natural trajectories use the following system prompt in all three domains. In the system prompt, we instruct the agent to turn the user instruction into a short keyword query consisting of the entity type and its key attributes, rather than copying the whole instruction or the price constraint into the query. We also instruct it to select all results that satisfy the instruction, leaving the number of selected results up to the agent.
At each step, the agent writes its reasoning inside \texttt{<think>} tags and then outputs a single action inside \texttt{<action>} tags, following the reasoning--action--observation loop of ReAct \citep{yao2022react}. We do not include any few-shot demonstrations.
The interaction is formatted as a multi-turn chat. The prompt is given as the system message, each environment observation as a user turn, and each model response as an assistant turn.

\begin{tcolorbox}[title={System prompt for natural trajectories},breakable,colback=white,colframe=black!35]
\begin{Verbatim}[fontsize=\footnotesize,breaklines=true,breakanywhere=true]
You are an autonomous web-search assistant. You are given an instruction describing what a user wants. Your job is to search the web and curate the set of results worth presenting to the user.

# How it works
Each turn you are shown the current page as plain text. Clickable elements appear inside square brackets, e.g. [ Search ], [ 3 ], [ Back to Search ].

# Actions
- search[<query>]: available only on the search page (when [ Search ] is shown). Write a concise, natural keyword query derived from the instruction (entity type plus key attributes). Do NOT put the whole instruction or the price constraint into the query.
- select[<n1,n2,...>]: available only on the results page. Provide the numbers (1-10) of the entities you would present to the user — every result that satisfies the instruction. Pick as many or as few as you judge appropriate (e.g. select[2] or select[1,4,7]). Submitting ends the task. To search again instead, write select[Back to Search].

# Flow
1. On the search page, issue one search[...] for the desired entity.
2. From the results, read the titles, urls and snippets, then decide which ones are worth showing the user.
3. Submit them with select[<n1,n2,...>], containing the numbers of the entities to present to the user. There is no purchase step — submitting finishes the task.

# Output format (STRICT)
First think step by step inside <think> </think> tags, then present the admissible action inside <action> </action> tags. Your return format should be:
<think>(your reasoning for this step)</think>
<action>(the action you chose in this step)</action>

Don't generate any other words after </action>.
\end{Verbatim}
\end{tcolorbox}

The prompt is supplied as a system message. The initial user message has the following form, with \texttt{\{instruction\}} replaced by the benchmark request:
\begin{Verbatim}[fontsize=\footnotesize,breaklines=true,breakanywhere=true]
Observation:
Instruction:
{instruction}

[ Search ]
\end{Verbatim}
After each model response, its full text is appended as an assistant message, followed by a user message containing \texttt{Observation:} and the environment's response. The accumulated messages are retained for subsequent turns. Search observations list numbered items with their title, content snippet, and URL, followed by \texttt{[ Back to Search ]}. The recorded configurations enable reasoning and multi-turn chat, retain URLs and full returned snippets, and do not shuffle the result order.
For deterministic decoding, we set the temperature to $0$ and the seed to $42$ where supported, except for GPT-5.6-Luna, which requires a temperature of $1$. However, these settings do not guarantee deterministic outputs from API-served models, whose underlying inference infrastructure and provider routing may remain outside our control.

\section{Evaluation Set Details}
\label{app:data-setting}

Table~\ref{tab:eval-sets} summarizes the evaluation sets. For Shopping, we use the first 1,500 requests from a shuffled collection of 12,087 WebShop goals. For Scholar, we use all 2,536 ScholarGym queries.

For Accommodation, we use all 214 HotelQuEST requests and add 572 location-substituted variants. To construct these variants, we first select 143 requests for which the target location can be replaced while keeping the rest of the request coherent and applicable. For each selected request, we generate four variants by replacing the original target location with another plausible location while leaving all other requirements unchanged. For example, ``budget-friendly hotel in Madrid'' is augmented to ``budget-friendly hotel in Toronto,'' and ``Looking for luxury, high-end hotel in Miami for our honeymoon'' is augmented to the same request targeting Lisbon. This yields $143 \times 4 = 572$ variants and 786 Accommodation requests in total.

Other than the WebShop subsampling and HotelQuEST augmentation described above, we apply no additional filtering or rewriting. Overall, the three evaluation sets contain 4,822 user requests.

\begin{table}[h]
\centering
\caption{Composition of the evaluation request sets across the three domains.}
\vspace{\baselineskip}
\label{tab:eval-sets}
\begin{tabular}{llrl}
\toprule
Domain & Benchmark & \#Requests & Construction \\
\midrule
Shopping & WebShop & 1,500 & First 1,500 from shuffled goals \\
Accommodation & HotelQuEST & 786 & All 214 + 572 location-substituted variants \\
Scholar & ScholarGym & 2,536 & All queries \\
\midrule
Total & & 4,822 & \\
\bottomrule
\end{tabular}
\end{table}

\section{Measuring User-Request Satisfaction}
\label{app:judge}

\paragraph{Request Requirements.}
For Shopping, we use the structured requirements already provided by WebShop, including product attributes, options, and price constraints. For Accommodation and Scholar, an LLM extracts verifiable requirements from each user request using the domain-specific prompts below. All extracted requirements in these two domains were reviewed by human judges. 

\paragraph{Validation of the LLM Judge.}
Search results retrieved through the live search API do not come with ground-truth labels indicating which requirements of the user's request they satisfy. We therefore use an LLM judge (Qwen3.8-27B) to assess each requirement separately from the result's title and content. To assess the reliability of these judgments, we compare them with human annotations on a sample of agent-selected results from each domain: 375 results for Shopping, 375 for Accommodation, and 100 for Scholar. For each request--result pair in this sample, two human annotators and the LLM judge assess whether each requirement is satisfied. Requirement judgments are binary, except that the Shopping price judgment also permits an unknown label when price information is unavailable. We measure agreement at the requirement level using simple agreement, Krippendorff's $\alpha$~\citep{hayes2007answering}, and Gwet's AC1~\citep{gwet2008computing}. Human--human agreement is computed between the two annotators assigned to each item. For human--LLM agreement, we pair each LLM judgment with each annotator's judgment and pool the resulting pairs within each domain before computing the agreement metrics.

\begin{wraptable}{r}{0.49\textwidth}
  \centering
  \caption{Feature-level human and judge agreement.   H--H denotes human--human agreement. H--L denotes pooled
  agreement between Qwen3.8 and each human annotation.}
  \vspace{\baselineskip}
  \label{tab:judge-feature-agreement}
  \vspace{0.5em}
  \resizebox{\linewidth}{!}{%
    \begin{tabular}{lrr*{3}{rr}}
      \toprule
      & & &
      \multicolumn{2}{c}{Agreement} &
      \multicolumn{2}{c}{Kripp.\ $\alpha$} &
      \multicolumn{2}{c}{Gwet's AC1} \\
      \cmidrule(lr){4-5}
      \cmidrule(lr){6-7}
      \cmidrule(lr){8-9}
      Domain & \# Items & \# Features
      & H--H & H--L
      & H--H & H--L
      & H--H & H--L \\
      \midrule
      Shopping
      & 375 & 1,265
      & .86 & .88
      & .72 & .75
      & .72 & .76 \\

      Accommodation
      & 375 & 1,198
      & .86 & .89
      & .69 & .74
      & .75 & .80 \\

      Scholar
      & 100 & 272
      & .85 & .85
      & .67 & .65
      & .72 & .73 \\
      \bottomrule
    \end{tabular}%
  }

\end{wraptable}

Across the three domains, human--LLM agreement broadly surpasses human--human agreement across all three metrics (Table~\ref{tab:judge-feature-agreement}). Human--LLM simple agreement ranged from 84.74\% to 88.61\%, Krippendorff's $\alpha$ from 0.647 to 0.750, and Gwet's AC1 from 0.732 to 0.798.

\paragraph{Requirement Extraction Prompts.}
The following system and user prompts are used to extract requirements for Accommodation and Scholar. Shopping requires no extraction prompt because its structured requirements are supplied by WebShop benchmark. Braced placeholders denote values inserted for each request; doubled braces used for escaping in the Python templates are displayed as literal braces.

\begin{tcolorbox}[title={Accommodation requirement extraction: system prompt},breakable,colback=white,colframe=black!35]
\begin{Verbatim}[fontsize=\footnotesize,breaklines=true,breakanywhere=true]
You extract the checkable requirements from a hotel/accommodation search query. Return each distinct, verifiable condition the user wants a result to satisfy. Avoid subjective wording (e.g. good, nice, highly); do NOT put such subjective expressions in the conditions. Avoid superlative wording (e.g. cheapest, best); state the condition in a concrete, checkable form instead.

Rules:
- One condition per requirement. Before returning, merge any two conditions that
  restate the same fact, differ only in wording, or where one contains the other.
  Never split one requirement across several conditions.
- Keep alternatives in ONE condition: "Prague or Budapest" -> "in Prague or
  Budapest". Never split a list of acceptable options into separate
  conditions.
- Conditions describe the accommodation, not the traveler. Emit the property the
  traveler's situation implies ("not afraid to pay" -> "upscale hotel"). Omit
  facts about the traveler themselves: age, motivation, who they travel with,
  how they get around, how long they stay.
- Mark exclusions with a prefix: "exclude: Isla Mujeres".
- Strip subjective adjectives and keep only the checkable noun: "good breakfast"
  -> "breakfast", "beautiful sea view" -> "sea view", "comfortable beds" ->
  "beds". If nothing checkable remains after stripping (e.g. "good location",
  "nice hotel", "clean place", "quiet"), drop the condition entirely.
- Do not drop any requirement the query states, and do not add preferences the
  query does not state or directly imply.

Respond with a single JSON object.
\end{Verbatim}
\end{tcolorbox}

\begin{tcolorbox}[title={Accommodation requirement extraction: user prompt},breakable,colback=white,colframe=black!35]
\begin{Verbatim}[fontsize=\footnotesize,breaklines=true,breakanywhere=true]
Query: {query}

Return JSON: {"conditions": ["<atomic condition>", ...]}
- Each condition is a short noun phrase a hotel result can be checked against.
- Example query: "cheapest hotel near the beach in Nice or Cannes for a
  family, not in the old town"
  -> {"conditions": ["low price", "near the beach", "in Nice or Cannes",
      "family rooms", "exclude: old town"]}
- Example query: "nice quiet hotel in Rome with good breakfast and a
  beautiful view"
  -> {"conditions": ["in Rome", "breakfast", "view"]}
Only output the JSON object.
\end{Verbatim}
\end{tcolorbox}

\begin{tcolorbox}[title={Scholar requirement extraction: system prompt},breakable,colback=white,colframe=black!35]
\begin{Verbatim}[fontsize=\footnotesize,breaklines=true,breakanywhere=true]
You extract the checkable requirements from an academic literature-search query. Each condition is one property a correct paper must have.

Rules:
- Delete any condition that is already part of another condition you return.
- "transferring facial expressions" and "facial expression transfer" are the
  same condition. Keep one.
- Every condition must make sense alone. Never return "based on", "using",
  "for", "from gloss sequences".
- Use only words the query uses. If it says "Deep Q-learning", do not write
  "DQN".
- Copy names exactly: GLaM, CLIP, WikiTQ, DataComp, BERT.
- Never split an "or". "detect or calibrate bias" is ONE condition, not two.
- "like A, B and C" lists examples, not requirements. Keep them in ONE
  condition and join them with "or": "A, B or C".
- Delete "first", "initially", "novel", "state-of-the-art", "well-known" - a
  title cannot prove them. "which paper FIRST proposed X" -> "proposes X".
- "without CAD models" becomes "exclude: CAD models". Never "CAD models".
- Write "is a dataset" / "is a benchmark" / "is a survey" only if the query
  asks for one. Never write "is a paper" - every answer is a paper already.
- Match the query: no requirement missing, none invented.

Respond with a single JSON object.
\end{Verbatim}
\end{tcolorbox}

\begin{tcolorbox}[title={Scholar requirement extraction: user prompt},breakable,colback=white,colframe=black!35]
\begin{Verbatim}[fontsize=\footnotesize,breaklines=true,breakanywhere=true]
Query: {query}

Return JSON: {"conditions": ["<condition>", ...]}

- "Which work pushes the limit of model quantization in BERT models by
  introducing a ternary network?"
  -> ["model quantization", "in BERT models", "ternary network"]

- "Could you provide me a study about generating sign pose sequences from gloss
  sequences by employing VQ-VAE?"
  WRONG -> ["generating sign pose sequences", "from gloss sequences",
            "employing VQ-VAE", "sign pose sequences", "gloss sequences"]
           (last two repeat the first; "from gloss sequences" is a fragment)
  RIGHT -> ["generating sign pose sequences from gloss sequences", "VQ-VAE"]

- "Could you provide me some works about fine-tuning LLMs to better respond to
  visual instructions?"
  WRONG -> ["fine-tuning LLMs", "visual instructions", "response to visual
            instructions", "visual instruction tuning", "multimodal instruction
            tuning", "image or vision input"]
           ("multimodal", "image or vision input" are not in the query)
  RIGHT -> ["fine-tuning LLMs", "responding to visual instructions"]

- "Could you suggest a dataset containing SQL annotations over WikiTQ, without
  depending on entity linking?"
  -> ["is a dataset", "SQL annotations", "over WikiTQ",
      "exclude: entity linking"]

- "Which paper first proposed multi-channel equivariant graph networks?"
  -> ["multi-channel equivariant graph networks"]

- "What papers proposed text data augmentation techniques like synonym
  replacement, positional swaps and back translation?"
  -> ["text data augmentation",
      "synonym replacement, positional swaps or back translation"]

Only output the JSON object.
\end{Verbatim}
\end{tcolorbox}

\paragraph{Requirement Satisfaction Prompts.}
The following domain-specific system and user prompts are used to judge each search result. The placeholders contain the user request, its structured or extracted requirements, and the result's title and content snippet. The result URL is not included in these prompt templates. For Scholar, we use the requirement-based rubric prompt, which assesses whether the paper covers each requested condition.

\begin{tcolorbox}[title={Shopping satisfaction judgment: system prompt},breakable,colback=white,colframe=black!35]
\begin{Verbatim}[fontsize=\footnotesize,breaklines=true,breakanywhere=true]
You are a strict evaluator. Given a shopping instruction's structured constraints and ONE web search result (title + snippet), decide whether the result satisfies each constraint based ONLY on evidence in the title/snippet. Do not assume unstated facts. Respond with a single JSON object.
\end{Verbatim}
\end{tcolorbox}

\begin{tcolorbox}[title={Shopping satisfaction judgment: user prompt},breakable,colback=white,colframe=black!35]
\begin{Verbatim}[fontsize=\footnotesize,breaklines=true,breakanywhere=true]
Instruction: {instruction}

Target product type (reference): {ptype}
Required attributes: {attrs}
Required options: {opts}
Price constraint: {price}

Search result:
  title: {title}
  snippet: {content}

Return JSON with EXACTLY these keys:
- "type_match": true/false  (is this the same kind of product as the target type?)
- "attributes": object mapping each required attribute -> true/false (evidenced in title/snippet?)
- "options": object mapping each required option -> true/false
- "price_ok": "yes" | "no" | "unknown"  (yes only if a price <= the limit is shown; no if a shown price exceeds it; unknown if no price shown or no limit)
- "reason": short string
Only output the JSON object.
\end{Verbatim}
\end{tcolorbox}

\begin{tcolorbox}[title={Accommodation satisfaction judgment: system prompt},breakable,colback=white,colframe=black!35]
\begin{Verbatim}[fontsize=\footnotesize,breaklines=true,breakanywhere=true]
You are a strict evaluator for hotel/accommodation search. Given the user's conditions and ONE search result (title + snippet), decide whether the result satisfies each condition based ONLY on evidence in the title/snippet. Do not assume unstated facts. Respond with a single JSON object.

Apply these rules:
- If the result lists several properties, a condition is satisfied when at least one listed property satisfies it.
- Evidence may come from the title or the snippet; either alone is enough.
- Quoted guest reviews and numeric ratings count as evidence, but only for the attribute the quote names.
- Proximity: if a distance or walking time is given, treat within 1.5 km or a 20-minute walk as satisfied; otherwise sharing the same named district, station or landmark counts as satisfied.
- A condition prefixed "exclude:" is satisfied by default; mark it false only if the snippet shows the result IS the excluded thing.
\end{Verbatim}
\end{tcolorbox}

\begin{tcolorbox}[title={Accommodation satisfaction judgment: user prompt},breakable,colback=white,colframe=black!35]
\begin{Verbatim}[fontsize=\footnotesize,breaklines=true,breakanywhere=true]
Instruction: {instruction}

Conditions the selected hotel must satisfy: {conditions}

Search result:
  title: {title}
  snippet: {content}

Return JSON with EXACTLY these keys:
- "conditions": object mapping EACH condition (verbatim) -> true/false (evidenced in title/snippet?)
- "reason": short string
Only output the JSON object.
\end{Verbatim}
\end{tcolorbox}

\begin{tcolorbox}[title={Scholar satisfaction judgment: system prompt},breakable,colback=white,colframe=black!35]
\begin{Verbatim}[fontsize=\footnotesize,breaklines=true,breakanywhere=true]
You are a strict evaluator for academic literature search. You are given a user's request, the conditions extracted from it, and ONE search result — a paper's title and an excerpt of its page. For each condition, decide whether the paper COVERS it: does the title or excerpt show that this work actually addresses that topic, method, or property? Judge only from the text given; do not assume unstated facts. Respond with a single JSON object.

Apply these rules:
- Evidence may come from the title or the excerpt; either alone is enough.
- The excerpt is a fragment of the page, not the full paper. Coverage shown anywhere in it counts; the condition need not be the work's main contribution.
- Naming a method only as prior work, related work, or a baseline does NOT count as the work covering it, when the condition requires the work to use or propose it.
- If the result is a page listing many papers, a condition is covered only when ONE listed paper covers it on its own.
- A condition prefixed "exclude:" is covered by default; mark it false only if the excerpt shows the result IS or DOES the excluded thing.
\end{Verbatim}
\end{tcolorbox}

\begin{tcolorbox}[title={Scholar satisfaction judgment: user prompt},breakable,colback=white,colframe=black!35]
\begin{Verbatim}[fontsize=\footnotesize,breaklines=true,breakanywhere=true]
Request: {instruction}

Conditions the paper must satisfy: {conditions}

Search result:
  title: {title}
  snippet: {content}

Return JSON with EXACTLY these keys:
- "conditions": object mapping EACH condition (verbatim) -> true/false
- "reason": short string
Only output the JSON object.
\end{Verbatim}
\end{tcolorbox}

\section{Measuring Source Preference: Estimation Details}
\label{app:bt}

\paragraph{Matched Comparisons.}
The two items $i$ and $j$ of a requirement-matched pair in a list of length $L$ each occupy every position exactly once across the $L$ rotated runs. For each position $p$, the run that places $i$ at $p$ and the run that places $j$ at $p$ form one comparison, scored as $D=1$ if only $i$ is selected, $D=-1$ if only $j$ is selected, and $D=0$ (a tie) otherwise (\S\ref{sec:measurement}). We treat both outcomes with $D=0$, both selected and neither selected, as ties, since both yield a zero selection difference. We drop comparisons that use a run in which the agent selected nothing, since such a run carries no information about preference. Table~\ref{tab:cells} gives the number of requests, requirement-matched pairs, and comparisons for each model and domain.

\paragraph{Preference Score.}
We fit the Bradley--Terry model with Davidson ties described in \S\ref{sec:measurement}, with one tie parameter $\nu$ per model and domain. Sources with fewer than 15 distinct requirement-matched pairs are not rated individually; they are pooled into a single opponent that shares one rating, and comparisons between two pooled sources are dropped. We maximize the likelihood with a ridge penalty $\lambda \sum_s \beta_s^2$ ($\lambda = 0.005$) on all ratings, including the pooled opponent, but not on $\nu$. The penalty keeps the estimate finite for a source that wins or loses every comparison and otherwise has almost no effect. After fitting, we center the ratings so that they sum to zero over rated sources; the pooled opponent is shifted by the same constant but excluded from the mean. The preference score $\tau_s$ (Eq.~\ref{eq:tau-s}) equals the expected score of $s$ against a source with $\beta = 0$ (win 1, tie $\tfrac12$, loss 0), rescaled to $[-1, 1]$.

\paragraph{Model Fit.}
The model summarizes each source by a single rating. To check this summary, for every pair of rated sources, or of a rated source and the pooled opponent, that meet in at least one comparison, we compare the observed mean score with the score implied by the fitted model (Table~\ref{tab:modelcheck}). Across source pairs, the correlation between the two ranges from 0.54 to 0.76, and the mean absolute residual, weighted by the number of comparisons, ranges from 0.02 to 0.06 on the 0--1 score scale.

\paragraph{Source Classification.}
Comparisons from the same request share lists and runs and are therefore not independent. We resample requests with replacement \citep{cameron2008bootstrap}, keep all comparisons of each drawn request, and refit the model, including $\nu$ and the pooled opponent, on each of $B = 10{,}000$ bootstrap replicates, with the pooling fixed from the full data and the ratings re-centered in each replicate. Confidence intervals are the 2.5th and 97.5th percentiles of the replicates. The two-sided $p$-value for $\beta_s = 0$ is $2\min\{\Pr[\hat\beta_s^* \le 0], \Pr[\hat\beta_s^* \ge 0]\}$ over the replicates, capped at 1; when all replicates fall on one side of zero, $p < 2/B$. We use all replicates, including the few whose fit stopped before convergence (at most 230 in one model--domain combination). Within each model and domain, we control the false discovery rate at 5\% \citep{benjamini1995controlling} and label a significant source \srcpos{} if $\beta_s > 0$ and \srcneg{} if $\beta_s < 0$; all other rated sources are \srczero{}.

A source that appears in only a few requests can be absent from a replicate. Such a replicate contains none of its comparisons, so the penalty sets its rating to nearly zero and the replicate counts as evidence in neither direction. Because one list can yield several requirement-matched pairs, such sources can still pass the 15-pair threshold: 10\% of rated sources are absent from at least 10\% of replicates. Tests for these sources are therefore conservative; of the 1{,}376 sources absent from more than 5\% of replicates, 1{,}373 are labeled \srczero{}.

\paragraph{Sensitivity to the Penalty.}
Refitting with $\lambda \in \{0, 0.01, 0.1, 1\}$ instead of 0.005 moves no source between \srcpos{} and \srcneg{}, and within each model and domain the rank correlation of ratings with those at $\lambda = 0.005$ is at least 0.98. What changes are labels near the significance threshold, mostly for sources with fewer than 30 requirement-matched pairs and mostly from \srczero{} to \srcpos{} or \srcneg{}: 1\% to 6\% of sources change label for $\lambda$ between 0.01 and 1, and 15\% without a penalty, because a source absent from a replicate then keeps its full-data estimate instead of zero. Redrawing the bootstrap at $\lambda = 0.005$ changes 0.4\%. The order $\mathrm{Inv}_{-,+} < \mathrm{Inv}_{0,0} < \mathrm{Inv}_{+,-}$ holds for every $\lambda$ in the 35 of 36 model--domain cells where all three rates are defined.

\begin{table}[t]\centering\small\setlength{\tabcolsep}{5pt}
\caption{Data behind each model and domain: requests contributing at least one comparison, requirement-matched pairs, position comparisons after excluding runs with no selection, rated sources (at least 15 requirement-matched pairs), and how many of them are labeled \srcpos{}, \srcneg{}, and \srczero{}.}
\label{tab:cells}
\begin{tabular}{@{}lrrrrrrr@{}}\toprule
Model & Requests & Pairs & Comparisons & Rated & \srcpos{} & \srcneg{} & \srczero{} \\ \midrule
\multicolumn{8}{@{}l}{\textit{Shopping}} \\
GPT-5.4-nano & 1{,}373 & 9{,}857 & 86{,}362 & 110 & 5 & 17 & 88 \\
GPT-5.6-Luna & 1{,}340 & 11{,}594 & 105{,}532 & 106 & 12 & 32 & 62 \\
Gemini-3.7-Flash & 785 & 7{,}053 & 50{,}866 & 77 & 7 & 8 & 62 \\
GLM-5.3-Flash & 1{,}318 & 10{,}887 & 82{,}077 & 138 & 12 & 21 & 105 \\
Deepseek-v4-flash & 1{,}094 & 8{,}232 & 61{,}039 & 95 & 8 & 8 & 79 \\
Llama-4-Maverick & 1{,}466 & 12{,}266 & 100{,}158 & 175 & 10 & 11 & 154 \\
Llama-4-Scout & 1{,}358 & 10{,}784 & 82{,}594 & 119 & 6 & 8 & 105 \\
Llama-3.1-8B-It & 1{,}407 & 11{,}427 & 94{,}860 & 144 & 8 & 12 & 124 \\
Tulu-3-8B & 1{,}431 & 10{,}410 & 90{,}365 & 130 & 5 & 11 & 114 \\
Qwen3.5-27B & 1{,}037 & 8{,}062 & 57{,}287 & 95 & 6 & 6 & 83 \\
Qwen3-30B-A3B & 1{,}419 & 10{,}686 & 92{,}536 & 125 & 5 & 9 & 111 \\
Qwen2.5-32B & 1{,}410 & 12{,}425 & 104{,}971 & 173 & 3 & 14 & 156 \\
\midrule
\multicolumn{8}{@{}l}{\textit{Accommodation}} \\
GPT-5.4-nano & 737 & 12{,}858 & 123{,}549 & 198 & 3 & 18 & 177 \\
GPT-5.6-Luna & 761 & 12{,}892 & 121{,}244 & 209 & 4 & 8 & 197 \\
Gemini-3.7-Flash & 626 & 12{,}529 & 108{,}177 & 213 & 2 & 6 & 205 \\
GLM-5.3-Flash & 710 & 12{,}686 & 114{,}924 & 192 & 13 & 8 & 171 \\
Deepseek-v4-flash & 630 & 11{,}400 & 97{,}224 & 184 & 5 & 8 & 171 \\
Llama-4-Maverick & 728 & 14{,}149 & 126{,}912 & 239 & 11 & 22 & 206 \\
Llama-4-Scout & 701 & 13{,}035 & 113{,}507 & 216 & 13 & 21 & 182 \\
Llama-3.1-8B-It & 727 & 13{,}640 & 121{,}762 & 243 & 4 & 12 & 227 \\
Tulu-3-8B & 736 & 13{,}355 & 125{,}678 & 247 & 3 & 14 & 230 \\
Qwen3.5-27B & 627 & 11{,}474 & 102{,}108 & 191 & 6 & 11 & 174 \\
Qwen3-30B-A3B & 741 & 13{,}010 & 121{,}443 & 194 & 16 & 19 & 159 \\
Qwen2.5-32B & 761 & 14{,}667 & 138{,}297 & 242 & 6 & 15 & 221 \\
\midrule
\multicolumn{8}{@{}l}{\textit{Scholar}} \\
GPT-5.4-nano & 2{,}505 & 48{,}205 & 427{,}246 & 402 & 62 & 57 & 283 \\
GPT-5.6-Luna & 2{,}426 & 50{,}633 & 470{,}080 & 443 & 49 & 69 & 325 \\
Gemini-3.7-Flash & 1{,}980 & 40{,}047 & 319{,}547 & 331 & 44 & 44 & 243 \\
GLM-5.3-Flash & 2{,}222 & 43{,}588 & 343{,}147 & 355 & 44 & 26 & 285 \\
Deepseek-v4-flash & 1{,}814 & 37{,}344 & 288{,}551 & 316 & 42 & 18 & 256 \\
Llama-4-Maverick & 2{,}516 & 49{,}356 & 442{,}989 & 421 & 23 & 20 & 378 \\
Llama-4-Scout & 2{,}487 & 48{,}753 & 431{,}074 & 441 & 27 & 25 & 389 \\
Llama-3.1-8B-It & 2{,}415 & 44{,}087 & 364{,}136 & 409 & 32 & 22 & 355 \\
Tulu-3-8B & 2{,}485 & 44{,}168 & 393{,}756 & 414 & 24 & 25 & 365 \\
Qwen3.5-27B & 1{,}987 & 34{,}508 & 286{,}206 & 308 & 43 & 24 & 241 \\
Qwen3-30B-A3B & 2{,}509 & 45{,}193 & 406{,}737 & 389 & 27 & 21 & 341 \\
Qwen2.5-32B & 2{,}468 & 43{,}305 & 377{,}281 & 404 & 39 & 37 & 328 \\
\bottomrule\end{tabular}\end{table}

\begin{table}[t]\centering\small\setlength{\tabcolsep}{5pt}
\caption{Fit of the Bradley--Terry model. For each pair of rated sources, or of a rated source and the pooled opponent, the observed mean score is compared with the score the fitted model implies. $r$: correlation between the two across source pairs; MAR: mean absolute residual, weighted by the number of comparisons.}
\label{tab:modelcheck}
\begin{tabular}{@{}lcc@{\hspace{10pt}}cc@{\hspace{10pt}}cc@{}}\toprule
 & \multicolumn{2}{c}{Shopping} & \multicolumn{2}{c}{Accommodation} & \multicolumn{2}{c}{Scholar} \\
\cmidrule(lr){2-3}\cmidrule(lr){4-5}\cmidrule(lr){6-7}
Model & $r$ & MAR & $r$ & MAR & $r$ & MAR \\ \midrule
GPT-5.4-nano & 0.64 & 0.038 & 0.76 & 0.052 & 0.69 & 0.048 \\
GPT-5.6-Luna & 0.72 & 0.028 & 0.72 & 0.056 & 0.65 & 0.052 \\
Gemini-3.7-Flash & 0.66 & 0.028 & 0.67 & 0.055 & 0.67 & 0.052 \\
GLM-5.3-Flash & 0.74 & 0.048 & 0.72 & 0.051 & 0.64 & 0.050 \\
Deepseek-v4-flash & 0.74 & 0.041 & 0.70 & 0.052 & 0.62 & 0.049 \\
Llama-4-Maverick & 0.68 & 0.043 & 0.71 & 0.052 & 0.57 & 0.048 \\
Llama-4-Scout & 0.63 & 0.033 & 0.70 & 0.046 & 0.56 & 0.044 \\
Llama-3.1-8B-It & 0.61 & 0.031 & 0.67 & 0.034 & 0.54 & 0.031 \\
Tulu-3-8B & 0.62 & 0.024 & 0.73 & 0.033 & 0.60 & 0.033 \\
Qwen3.5-27B & 0.66 & 0.042 & 0.72 & 0.061 & 0.63 & 0.056 \\
Qwen3-30B-A3B & 0.60 & 0.037 & 0.68 & 0.054 & 0.57 & 0.046 \\
Qwen2.5-32B & 0.66 & 0.033 & 0.71 & 0.045 & 0.58 & 0.039 \\
\bottomrule\end{tabular}\end{table}

\FloatBarrier
\section{Source Preference in the Wild: Full Results}
\label{app:wild}
\begin{figure}[t]\centering
\includegraphics[width=0.45\linewidth]{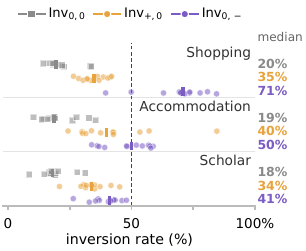}
\caption{Inversion rates for pairs in which exactly one item comes from a neutral source. Dot: one model; bar: median. Values in Table~\ref{tab:inversion-side}.}
\label{fig:inversion-side}
\end{figure}

Tables~\ref{tab:wild-top20-webshop}--\ref{tab:wild-top20-scholar} give, for each domain, the per-model values behind Fig.~\ref{fig:wild}. Table~\ref{tab:wild-disagree} lists the sources on which models disagree in direction. Table~\ref{tab:inversion-n} gives the values behind Fig.~\ref{fig:inversion}. Table~\ref{tab:inversion-side} gives the inversion rates for pairs in which exactly one item comes from a neutral source, plotted in Fig.~\ref{fig:inversion-side}.

\begin{table}[t]\centering\small
\definecolor{pos}{HTML}{E58484}\definecolor{neg}{HTML}{6D9EEB}
\newcommand{\pc}[2]{\cellcolor{pos!#1!white}$+#2$}\newcommand{\nc}[2]{\cellcolor{neg!#1!white}$-#2$}\newcommand{\zc}[1]{\textcolor{gray}{$#1$}}
\caption{Shopping: $\tau_s$ (percentage points) of the 20 sources that appear most often in the search results (the sources of Fig.~\ref{fig:wild}), by model. Share: fraction of all search results, pooled over models. Cells are colored when the model labels the source \srcpos{} (red) or \srcneg{} (blue); gray: \srczero{}. Maj.: the label held by a majority of the models that rate the source (--: none). Models: GPT = GPT-5.4-nano, Luna = GPT-5.6-Luna, Gem = Gemini-3.7-Flash, GLM = GLM-5.3-Flash, DS = Deepseek-v4-flash, Mav = Llama-4-Maverick, Sct = Llama-4-Scout, L3.1 = Llama-3.1-8B-It, Tulu = Tulu-3-8B, Q3.5 = Qwen3.5-27B, Q3 = Qwen3-30B-A3B, Q2.5 = Qwen2.5-32B.}
\label{tab:wild-top20-webshop}
\setlength{\tabcolsep}{3pt}\renewcommand{\arraystretch}{1.1}
\resizebox{\linewidth}{!}{%
\begin{tabular}{@{}rlrccccccccccccc@{}}\toprule
Rank & Source & Share & GPT & Luna & Gem & GLM & DS & Mav & Sct & L3.1 & Tulu & Q3.5 & Q3 & Q2.5 & Maj. \\ \midrule
1 & Amazon & 12.2\% & \nc{12}{6} & \pc{33}{16} & \pc{12}{6} & \pc{30}{15} & \pc{29}{14} & \pc{53}{27} & \pc{37}{19} & \pc{12}{6} & \pc{10}{5} & \pc{52}{26} & \pc{31}{16} & \pc{31}{15} & $+$ \\
2 & Walmart & 6.5\% & \pc{22}{11} & \pc{48}{24} & \pc{22}{11} & \pc{34}{17} & \pc{29}{15} & \pc{50}{25} & \pc{41}{21} & \pc{23}{12} & \pc{22}{11} & \pc{40}{20} & \pc{42}{21} & \pc{33}{17} & $+$ \\
3 & eBay & 5.1\% & \nc{36}{18} & \pc{45}{23} & \pc{23}{12} & \pc{31}{16} & \pc{17}{9} & \zc{-1} & \zc{+3} & \zc{+2} & \zc{+0} & \zc{+1} & \zc{+3} & \pc{15}{8} & -- \\
4 & Target & 2.7\% & \zc{+5} & \pc{17}{8} & \zc{-0} & \zc{+8} & \zc{+5} & \zc{+7} & \zc{+3} & \zc{+6} & \pc{24}{12} & \zc{+10} & \zc{+8} & \zc{+4} & $0$ \\
5 & YouTube & 2.6\% & \nc{59}{29} & \nc{24}{12} & \nc{45}{22} & \nc{70}{35} & \nc{70}{35} & \nc{54}{27} & \nc{27}{14} & \nc{42}{21} & \nc{44}{22} & \nc{60}{30} & \nc{43}{22} & \nc{33}{16} & $-$ \\
6 & Facebook & 1.9\% & \nc{62}{31} & \nc{27}{13} & \nc{42}{21} & \nc{73}{37} & \nc{70}{35} & \nc{47}{24} & \nc{37}{19} & \nc{46}{23} & \nc{37}{19} & \nc{57}{28} & \nc{45}{23} & \nc{38}{19} & $-$ \\
7 & Pinterest & 1.8\% & \nc{57}{28} & \nc{23}{12} & \nc{37}{19} & \nc{75}{38} & \nc{90}{45} & \nc{55}{28} & \nc{40}{20} & \nc{49}{24} & \nc{26}{13} & \nc{48}{24} & \nc{55}{27} & \nc{42}{21} & $-$ \\
8 & Reddit & 1.8\% & \nc{60}{30} & \nc{26}{13} & \nc{45}{23} & \nc{74}{37} & \nc{40}{20} & \nc{49}{24} & \nc{40}{20} & \nc{44}{22} & \nc{36}{18} & \nc{66}{33} & \nc{46}{23} & \nc{43}{22} & $-$ \\
9 & Etsy & 1.7\% & \nc{28}{14} & \zc{+8} & \zc{-8} & \zc{+1} & \zc{-0} & \zc{-9} & \zc{+0} & \nc{18}{9} & \zc{+3} & \zc{-5} & \zc{-2} & \zc{+2} & $0$ \\
10 & Home Depot & 1.6\% & \pc{21}{10} & \zc{+4} & \zc{+4} & \zc{+9} & \zc{+9} & \pc{25}{13} & \zc{+4} & \pc{18}{9} & \pc{17}{8} & \zc{+6} & \zc{+6} & \zc{+1} & $0$ \\
11 & Business Insider & 1.1\% & \zc{-45} & \nc{25}{12} & -- & -- & -- & -- & -- & -- & -- & -- & \zc{-4} & -- & $0$ \\
12 & Best Buy & 1.0\% & \zc{+1} & \zc{+16} & \zc{+3} & \zc{+8} & \zc{+5} & \zc{+7} & \zc{+3} & \zc{+5} & \zc{+4} & \zc{+11} & \zc{+5} & \zc{-0} & $0$ \\
13 & Yahoo & 0.9\% & \nc{57}{28} & \nc{20}{10} & -- & -- & -- & -- & -- & -- & \zc{-4} & -- & -- & -- & $-$ \\
14 & Wayfair & 0.9\% & \zc{-3} & \zc{+9} & \zc{+7} & \zc{+5} & \zc{+1} & \pc{35}{17} & \pc{29}{15} & \pc{31}{16} & \zc{+7} & \pc{36}{18} & \zc{+7} & \zc{-0} & $0$ \\
15 & Bed Bath \& Beyond & 0.8\% & \zc{-4} & \zc{+14} & \zc{+8} & \zc{+2} & \zc{+3} & \zc{-6} & \zc{+0} & \zc{+4} & \zc{+6} & \zc{+8} & \zc{+2} & \zc{-4} & $0$ \\
16 & Instagram & 0.6\% & \nc{48}{24} & \nc{36}{18} & \nc{32}{16} & \nc{74}{37} & \nc{70}{35} & \nc{50}{25} & \nc{35}{18} & \nc{61}{31} & \nc{30}{15} & \nc{50}{25} & \nc{43}{22} & \nc{34}{17} & $-$ \\
17 & Lowe's & 0.6\% & \zc{+7} & \pc{44}{22} & \zc{+4} & \pc{46}{23} & \zc{+12} & \zc{-3} & \zc{+8} & \pc{26}{13} & \zc{+10} & \zc{-4} & \zc{+8} & \zc{-5} & $0$ \\
18 & Macy's & 0.6\% & \zc{+1} & \zc{+4} & \zc{-0} & \zc{+1} & \zc{+3} & \zc{-6} & \zc{+6} & \zc{+0} & \zc{+7} & \zc{+3} & \zc{-6} & \zc{-0} & $0$ \\
19 & Kohl's & 0.5\% & \pc{24}{12} & \pc{45}{22} & \zc{+21} & \pc{53}{26} & \zc{+13} & \zc{-4} & \zc{+9} & \zc{+8} & \zc{+2} & \zc{+4} & \pc{38}{19} & \zc{+14} & $0$ \\
20 & New York Times & 0.5\% & \zc{-6} & \zc{-2} & \nc{34}{17} & \zc{-7} & \zc{+3} & \zc{-7} & \zc{+2} & \nc{27}{13} & \nc{23}{11} & \zc{-13} & \zc{-6} & \zc{-6} & $0$ \\
\bottomrule\end{tabular}}\end{table}

\begin{table}[t]\centering\small
\definecolor{pos}{HTML}{E58484}\definecolor{neg}{HTML}{6D9EEB}
\newcommand{\pc}[2]{\cellcolor{pos!#1!white}$+#2$}\newcommand{\nc}[2]{\cellcolor{neg!#1!white}$-#2$}\newcommand{\zc}[1]{\textcolor{gray}{$#1$}}
\caption{Accommodation: as Table~\ref{tab:wild-top20-webshop}.}
\label{tab:wild-top20-accommodation}
\setlength{\tabcolsep}{3pt}\renewcommand{\arraystretch}{1.1}
\resizebox{\linewidth}{!}{%
\begin{tabular}{@{}rlrccccccccccccc@{}}\toprule
Rank & Source & Share & GPT & Luna & Gem & GLM & DS & Mav & Sct & L3.1 & Tulu & Q3.5 & Q3 & Q2.5 & Maj. \\ \midrule
1 & Tripadvisor & 8.9\% & \nc{28}{14} & \nc{24}{12} & \zc{-4} & \zc{+5} & \zc{-3} & \nc{22}{11} & \nc{25}{13} & \nc{16}{8} & \nc{16}{8} & \zc{+2} & \nc{12}{6} & \nc{19}{10} & $-$ \\
2 & Facebook & 6.5\% & \nc{100}{52} & \nc{83}{42} & \nc{87}{43} & \nc{100}{51} & \nc{98}{49} & \nc{86}{43} & \nc{61}{31} & \nc{38}{19} & \nc{57}{28} & \nc{92}{46} & \nc{70}{35} & \nc{65}{32} & $-$ \\
3 & Expedia & 4.3\% & \zc{-5} & \zc{-9} & \zc{-6} & \zc{-7} & \zc{-8} & \nc{43}{21} & \nc{35}{18} & \zc{-3} & \nc{18}{9} & \nc{36}{18} & \nc{36}{18} & \nc{16}{8} & -- \\
4 & Booking.com & 3.5\% & \pc{22}{11} & \pc{31}{15} & \zc{+6} & \pc{31}{15} & \pc{27}{13} & \pc{19}{10} & \zc{+7} & \pc{14}{7} & \pc{17}{9} & \pc{43}{21} & \pc{17}{8} & \pc{17}{8} & $+$ \\
5 & Reddit & 3.0\% & \nc{100}{51} & \nc{94}{47} & \nc{79}{39} & \nc{53}{27} & \nc{50}{25} & \nc{88}{44} & \nc{71}{35} & \nc{48}{24} & \nc{62}{31} & \nc{83}{41} & \nc{69}{35} & \nc{70}{35} & $-$ \\
6 & Hotels.com & 2.2\% & \zc{-7} & \zc{-4} & \zc{-6} & \zc{+0} & \zc{-3} & \nc{41}{20} & \nc{33}{17} & \nc{14}{7} & \nc{14}{7} & \zc{-10} & \nc{39}{20} & \nc{27}{13} & -- \\
7 & YouTube & 2.0\% & \nc{93}{46} & \nc{85}{42} & \nc{67}{34} & \nc{78}{39} & \nc{82}{41} & \nc{85}{42} & \nc{43}{21} & \nc{16}{8} & \nc{47}{24} & \nc{90}{45} & \nc{55}{28} & \nc{42}{21} & $-$ \\
8 & Instagram & 1.8\% & \nc{99}{50} & \nc{49}{24} & \nc{67}{34} & \nc{100}{52} & \nc{87}{44} & \nc{62}{31} & \nc{55}{27} & \nc{38}{19} & \nc{51}{26} & \nc{80}{40} & \nc{38}{19} & \nc{51}{26} & $-$ \\
9 & Marriott & 1.3\% & \zc{-6} & \zc{+4} & \zc{-1} & \zc{+0} & \zc{-1} & \nc{34}{17} & \zc{-2} & \zc{+0} & \zc{+0} & \zc{-11} & \zc{-4} & \zc{-5} & $0$ \\
10 & Condé Nast Traveler & 1.2\% & \zc{+6} & \zc{+5} & \zc{-8} & \zc{-3} & \zc{+1} & \zc{-9} & \nc{24}{12} & \zc{-4} & \zc{-2} & \zc{-0} & \zc{-2} & \zc{-2} & $0$ \\
11 & Trip.com & 1.2\% & \nc{56}{28} & \zc{-4} & \zc{-2} & \zc{+1} & \zc{-2} & \zc{-6} & \nc{23}{12} & \zc{+3} & \zc{-5} & \zc{-3} & \nc{31}{16} & \zc{-10} & $0$ \\
12 & Agoda & 1.2\% & \nc{37}{19} & \zc{-13} & \zc{-9} & \zc{-11} & \nc{42}{21} & \nc{36}{18} & \nc{45}{22} & \zc{-2} & \zc{-11} & \nc{56}{28} & \nc{53}{27} & \zc{-7} & -- \\
13 & Travelmyth (.com) & 1.0\% & \zc{-10} & \zc{-13} & \zc{+1} & \zc{+5} & \zc{+5} & \zc{-7} & \zc{-5} & \zc{-9} & \zc{-4} & \zc{-19} & \zc{-9} & \zc{-9} & $0$ \\
14 & Travelmyth (.co.uk) & 0.9\% & \zc{+14} & \zc{-19} & \zc{-27} & \zc{-1} & \zc{-13} & \zc{-31} & \zc{-12} & \zc{-12} & \zc{-7} & \zc{-1} & \zc{-22} & \zc{-9} & $0$ \\
15 & KAYAK & 0.9\% & \zc{+10} & \zc{+17} & \zc{+9} & \zc{+6} & \pc{34}{17} & \zc{-6} & \zc{0} & \zc{-2} & \zc{-3} & \zc{+6} & \zc{+8} & \zc{+8} & $0$ \\
16 & Travelocity & 0.8\% & \zc{-19} & \zc{-16} & \zc{-10} & \zc{-15} & \zc{-9} & \nc{68}{34} & \nc{45}{22} & \zc{-7} & \nc{21}{10} & \nc{73}{36} & \nc{61}{31} & \nc{25}{12} & -- \\
17 & trivago & 0.8\% & \zc{-14} & \zc{+2} & \zc{-5} & \zc{-3} & \zc{-2} & \nc{26}{13} & \nc{31}{15} & \zc{0} & \zc{-1} & \zc{+4} & \zc{-4} & \zc{-10} & $0$ \\
18 & Rick Steves & 0.7\% & \nc{88}{44} & \nc{83}{42} & \nc{60}{30} & \nc{53}{26} & \nc{48}{24} & \nc{72}{36} & \nc{63}{32} & \nc{35}{18} & \nc{44}{22} & \zc{-19} & \nc{54}{27} & \nc{65}{32} & $-$ \\
19 & Hilton & 0.5\% & \pc{61}{31} & \zc{+6} & \zc{-1} & \zc{+9} & \zc{-6} & \zc{-5} & \zc{+14} & \zc{+2} & \zc{+12} & \zc{+22} & \zc{-7} & \zc{+6} & $0$ \\
20 & IHG & 0.5\% & \zc{-21} & \zc{+5} & \zc{-8} & \zc{-5} & \zc{-23} & \zc{-11} & \zc{-11} & \zc{+1} & \zc{+12} & \zc{-1} & \zc{-15} & \zc{-5} & $0$ \\
\bottomrule\end{tabular}}\end{table}

\begin{table}[t]\centering\small
\definecolor{pos}{HTML}{E58484}\definecolor{neg}{HTML}{6D9EEB}
\newcommand{\pc}[2]{\cellcolor{pos!#1!white}$+#2$}\newcommand{\nc}[2]{\cellcolor{neg!#1!white}$-#2$}\newcommand{\zc}[1]{\textcolor{gray}{$#1$}}
\caption{Scholar: as Table~\ref{tab:wild-top20-webshop}.}
\label{tab:wild-top20-scholar}
\setlength{\tabcolsep}{3pt}\renewcommand{\arraystretch}{1.1}
\resizebox{\linewidth}{!}{%
\begin{tabular}{@{}rlrccccccccccccc@{}}\toprule
Rank & Source & Share & GPT & Luna & Gem & GLM & DS & Mav & Sct & L3.1 & Tulu & Q3.5 & Q3 & Q2.5 & Maj. \\ \midrule
1 & arXiv & 11.5\% & \pc{67}{34} & \pc{57}{28} & \pc{58}{29} & \pc{65}{33} & \pc{53}{27} & \pc{51}{25} & \pc{56}{28} & \pc{31}{16} & \pc{39}{19} & \pc{65}{32} & \pc{47}{23} & \pc{46}{23} & $+$ \\
2 & ScienceDirect & 4.2\% & \pc{22}{11} & \pc{36}{18} & \pc{29}{15} & \pc{34}{17} & \pc{26}{13} & \zc{+2} & \zc{-1} & \pc{6}{3} & \zc{+1} & \pc{28}{14} & \pc{15}{7} & \pc{16}{8} & $+$ \\
3 & OpenReview & 3.9\% & \pc{57}{29} & \pc{41}{21} & \pc{43}{21} & \pc{42}{21} & \pc{29}{14} & \zc{+1} & \zc{+3} & \pc{11}{5} & \pc{6}{3} & \pc{36}{18} & \zc{+3} & \pc{15}{8} & $+$ \\
4 & Medium & 3.9\% & \nc{47}{24} & \nc{45}{23} & \nc{52}{26} & \nc{42}{21} & \nc{41}{21} & \nc{22}{11} & \nc{22}{11} & \nc{10}{5} & \nc{11}{5} & \nc{48}{24} & \nc{26}{13} & \nc{13}{7} & $-$ \\
5 & GitHub & 3.6\% & \zc{-3} & \nc{15}{7} & \nc{19}{10} & \pc{27}{14} & \pc{38}{19} & \zc{+0} & \zc{+1} & \nc{12}{6} & \nc{15}{7} & \pc{24}{12} & \zc{+1} & \pc{8}{4} & -- \\
6 & GitHub Pages & 2.9\% & \pc{34}{17} & \pc{15}{8} & \pc{16}{8} & \pc{25}{13} & \pc{22}{11} & \pc{24}{12} & \pc{26}{13} & \pc{16}{8} & \pc{16}{8} & \pc{31}{16} & \pc{21}{11} & \pc{22}{11} & $+$ \\
7 & NIH & 2.8\% & \pc{39}{20} & \pc{41}{21} & \pc{32}{16} & \pc{32}{16} & \pc{33}{16} & \pc{22}{11} & \pc{13}{6} & \pc{11}{5} & \zc{+2} & \pc{33}{17} & \pc{20}{10} & \pc{16}{8} & $+$ \\
8 & ResearchGate & 2.8\% & \nc{22}{11} & \nc{10}{5} & \pc{37}{19} & \zc{+2} & \zc{-4} & \zc{-0} & \zc{-3} & \zc{+2} & \zc{+1} & \zc{+1} & \zc{-2} & \pc{8}{4} & $0$ \\
9 & ACM & 2.6\% & \pc{39}{20} & \pc{47}{23} & \pc{49}{24} & \pc{45}{23} & \pc{29}{14} & \zc{0} & \zc{+1} & \pc{16}{8} & \pc{11}{5} & \pc{32}{16} & \zc{+3} & \pc{15}{7} & $+$ \\
10 & NeurIPS & 2.2\% & \pc{67}{34} & \pc{68}{34} & \pc{40}{20} & \pc{55}{27} & \pc{41}{20} & \pc{41}{20} & \pc{45}{23} & \pc{32}{16} & \pc{26}{13} & \pc{57}{29} & \pc{46}{23} & \pc{36}{18} & $+$ \\
11 & ACL Anthology & 2.2\% & \pc{93}{47} & \pc{85}{42} & \pc{77}{38} & \pc{71}{36} & \pc{59}{30} & \pc{47}{24} & \pc{50}{25} & \pc{37}{18} & \pc{32}{16} & \pc{76}{38} & \pc{48}{24} & \pc{45}{22} & $+$ \\
12 & Emergent Mind & 2.2\% & \nc{33}{17} & \nc{44}{22} & \nc{30}{15} & \zc{+1} & \zc{-3} & \pc{26}{13} & \pc{12}{6} & \pc{9}{4} & \pc{9}{4} & \zc{-2} & \pc{20}{10} & \pc{19}{9} & -- \\
13 & Semantic Scholar & 2.0\% & \pc{14}{7} & \zc{-0} & \pc{54}{27} & \pc{27}{14} & \pc{21}{10} & \zc{+4} & \zc{+3} & \zc{+2} & \zc{+2} & \pc{26}{13} & \pc{12}{6} & \pc{25}{13} & $+$ \\
14 & Reddit & 2.0\% & \nc{42}{21} & \nc{41}{21} & \nc{67}{33} & \nc{61}{30} & \nc{67}{34} & \nc{59}{29} & \nc{51}{25} & \nc{44}{22} & \nc{57}{28} & \nc{59}{30} & \nc{77}{38} & \nc{45}{22} & $-$ \\
15 & YouTube & 1.6\% & \nc{35}{18} & \nc{35}{18} & \nc{65}{33} & \nc{37}{18} & \nc{63}{31} & \nc{53}{26} & \nc{37}{19} & \nc{20}{10} & \nc{34}{17} & \nc{67}{34} & \nc{65}{32} & \nc{28}{14} & $-$ \\
16 & alphaXiv & 1.5\% & \pc{12}{6} & \zc{-4} & \pc{46}{23} & \pc{24}{12} & \pc{17}{8} & \pc{21}{11} & \pc{26}{13} & \pc{33}{17} & \pc{28}{14} & \pc{21}{10} & \pc{44}{22} & \pc{35}{18} & $+$ \\
17 & Springer & 1.4\% & \pc{36}{18} & \pc{35}{18} & \pc{43}{22} & \pc{50}{25} & \pc{36}{18} & \pc{37}{18} & \pc{20}{10} & \pc{14}{7} & \pc{11}{6} & \pc{38}{19} & \pc{29}{15} & \pc{24}{12} & $+$ \\
18 & MDPI & 1.4\% & \pc{37}{19} & \pc{45}{22} & \pc{48}{24} & \pc{49}{24} & \pc{43}{21} & \pc{35}{18} & \pc{18}{9} & \pc{24}{12} & \pc{15}{7} & \pc{46}{23} & \pc{32}{16} & \pc{24}{12} & $+$ \\
19 & IEEE & 1.3\% & \pc{54}{27} & \pc{50}{25} & \pc{56}{28} & \pc{52}{26} & \pc{33}{16} & \pc{19}{9} & \pc{18}{9} & \pc{25}{13} & \pc{11}{5} & \pc{48}{24} & \pc{21}{10} & \pc{25}{12} & $+$ \\
20 & PMLR & 1.3\% & \pc{83}{42} & \pc{91}{46} & \pc{67}{34} & \pc{71}{35} & \pc{54}{27} & \pc{56}{28} & \pc{56}{28} & \pc{32}{16} & \pc{40}{20} & \pc{77}{38} & \pc{41}{20} & \pc{36}{18} & $+$ \\
\bottomrule\end{tabular}}\end{table}

\begin{table}[t]\centering\footnotesize
\caption{Sources on which models disagree in direction: at least one model labels the source \srcpos{} and at least one labels it \srcneg{} (labels after FDR correction, \S\ref{sec:measurement}). Values in parentheses are $\tau_s$ in percentage points; lists of seven or more models give only the count. \srczero{}: number of models that rate the source but label it \srczero{}. Rank: rank by frequency in the search results of the domain, pooled over models. Model abbreviations as in Table~\ref{tab:wild-top20-webshop}.}
\label{tab:wild-disagree}
\setlength{\tabcolsep}{4pt}\renewcommand{\arraystretch}{1.1}
\begin{tabularx}{\linewidth}{@{}llrYYr@{}}\toprule
Domain & Source & Rank & \srcpos{} (models, $\tau_s$) & \srcneg{} (models, $\tau_s$) & \srczero{} \\ \midrule
Shopping & Amazon & 1 & 11 models & GPT ($-$6) & 0 \\
 & eBay & 3 & Luna (+23), Gem (+12), GLM (+16), DS (+9), Q2.5 (+8) & GPT ($-$18) & 6 \\
\midrule
Accommodation & Forbes Travel Guide & 34 & Q3 (+22) & Mav ($-$20) & 10 \\
\midrule
Scholar & GitHub & 5 & GLM (+14), DS (+19), Q3.5 (+12), Q2.5 (+4) & Luna ($-$7), Gem ($-$10), L3.1 ($-$6), Tulu ($-$7) & 4 \\
 & ResearchGate & 8 & Gem (+19), Q2.5 (+4) & GPT ($-$11), Luna ($-$5) & 8 \\
 & Emergent Mind & 12 & Mav (+13), Sct (+6), L3.1 (+4), Tulu (+4), Q3 (+10), Q2.5 (+9) & GPT ($-$17), Luna ($-$22), Gem ($-$15) & 3 \\
 & Liner & 30 & Sct (+14), L3.1 (+7), Q3 (+19), Q2.5 (+15) & GPT ($-$13), Luna ($-$14) & 6 \\
 & Moonlight & 44 & Q2.5 (+11) & GPT ($-$17), Luna ($-$19), Mav ($-$16) & 8 \\
 & mbrenndoerfer.com & 61 & Q3 (+20) & Luna ($-$19), GLM ($-$15) & 9 \\
 & aman.ai & 123 & Tulu (+21) & GPT ($-$31) & 10 \\
 & UC San Diego & 132 & Q3.5 (+34) & GPT ($-$11) & 10 \\
 & University of Michigan & 140 & Q2.5 (+29) & Q3 ($-$22) & 10 \\
 & TOPBOTS & 155 & DS (+27) & GPT ($-$22), Luna ($-$21) & 9 \\
 & eugeneyan.com & 163 & Sct (+21) & GPT ($-$37), Luna ($-$30) & 9 \\
 & Brown University & 182 & Gem (+23) & Luna ($-$30) & 6 \\
 & Learn Prompting & 199 & Sct (+29) & GPT ($-$39) & 10 \\
\bottomrule\end{tabularx}\end{table}

\begin{table}[t]\centering\small\setlength{\tabcolsep}{4pt}
\newcommand{\tbd}{\textcolor{gray}{tbd}}
\caption{Inversion rates (\%) of Fig.~\ref{fig:inversion}. In parentheses: request-level bootstrap 95\% confidence interval; number of comparisons with exactly one item selected. --: no pairs of this kind; Gemini-3.7-Flash labels only two Accommodation sources \srcpos{}, which yields no $(+,-)$ pairs there. Across the three rates, the number of such comparisons per cell has a median of about 2{,}000 and is below 100 only for two Gemini-3.7-Flash cells.}
\label{tab:inversion-n}
\resizebox{\linewidth}{!}{%
\begin{tabular}{@{}lccc@{\hspace{6pt}}ccc@{\hspace{6pt}}ccc@{}}\toprule
 & \multicolumn{3}{c}{Shopping} & \multicolumn{3}{c}{Accommodation} & \multicolumn{3}{c}{Scholar} \\
\cmidrule(lr){2-4}\cmidrule(lr){5-7}\cmidrule(lr){8-10}
Model & $\mathrm{Inv}_{-,+}$ & $\mathrm{Inv}_{0,0}$ & $\mathrm{Inv}_{+,-}$ & $\mathrm{Inv}_{-,+}$ & $\mathrm{Inv}_{0,0}$ & $\mathrm{Inv}_{+,-}$ & $\mathrm{Inv}_{-,+}$ & $\mathrm{Inv}_{0,0}$ & $\mathrm{Inv}_{+,-}$ \\ \midrule
GPT-5.4-nano & 3 {\scriptsize(2--5; 2{,}021)} & 20 {\scriptsize(15--25; 1{,}646)} & 54 {\scriptsize(46--62; 1{,}148)} & 7 {\scriptsize(3--12; 1{,}226)} & 26 {\scriptsize(21--31; 3{,}142)} & 72 {\scriptsize(56--84; 380)} & 1 {\scriptsize(1--1; 20{,}538)} & 19 {\scriptsize(14--25; 1{,}770)} & 68 {\scriptsize(64--73; 6{,}489)} \\
GPT-5.6-Luna & 0 {\scriptsize(0--1; 7{,}313)} & 18 {\scriptsize(7--33; 369)} & 92 {\scriptsize(85--97; 644)} & 3 {\scriptsize(1--6; 899)} & 14 {\scriptsize(10--18; 5{,}310)} & 69 {\scriptsize(52--83; 222)} & 2 {\scriptsize(2--3; 24{,}847)} & 20 {\scriptsize(16--25; 3{,}995)} & 69 {\scriptsize(65--73; 10{,}597)} \\
Gemini-3.7-Flash & 0 {\scriptsize(0--0; 926)} & 19 {\scriptsize(7--34; 447)} & 97 {\scriptsize(85--100; 39)} & 0 {\scriptsize(0--0; 44)} & 10 {\scriptsize(8--13; 8{,}288)} & -- & 2 {\scriptsize(1--3; 16{,}286)} & 9 {\scriptsize(5--13; 1{,}428)} & 48 {\scriptsize(42--53; 5{,}063)} \\
GLM-5.3-Flash & 1 {\scriptsize(0--1; 3{,}147)} & 22 {\scriptsize(16--29; 1{,}534)} & 71 {\scriptsize(55--88; 334)} & 1 {\scriptsize(0--2; 1{,}052)} & 16 {\scriptsize(13--19; 7{,}783)} & 85 {\scriptsize(70--96; 201)} & 1 {\scriptsize(1--2; 12{,}336)} & 15 {\scriptsize(11--18; 3{,}784)} & 63 {\scriptsize(56--69; 2{,}737)} \\
Deepseek-v4-flash & 1 {\scriptsize(0--1; 1{,}818)} & 18 {\scriptsize(14--23; 1{,}378)} & 77 {\scriptsize(63--89; 300)} & 2 {\scriptsize(1--5; 790)} & 19 {\scriptsize(15--23; 5{,}823)} & 50 {\scriptsize(30--75; 133)} & 1 {\scriptsize(1--2; 9{,}527)} & 17 {\scriptsize(14--21; 3{,}545)} & 56 {\scriptsize(49--64; 2{,}163)} \\
Llama-4-Maverick & 1 {\scriptsize(0--2; 2{,}202)} & 23 {\scriptsize(19--27; 2{,}523)} & 91 {\scriptsize(84--95; 338)} & 5 {\scriptsize(2--8; 1{,}712)} & 19 {\scriptsize(15--24; 1{,}963)} & 56 {\scriptsize(42--69; 450)} & 1 {\scriptsize(1--2; 12{,}842)} & 17 {\scriptsize(15--18; 19{,}191)} & 57 {\scriptsize(50--64; 2{,}438)} \\
Llama-4-Scout & 0 {\scriptsize(0--1; 1{,}413)} & 18 {\scriptsize(14--22; 2{,}145)} & 80 {\scriptsize(63--94; 235)} & 3 {\scriptsize(0--9; 419)} & 28 {\scriptsize(22--34; 1{,}993)} & 93 {\scriptsize(82--98; 192)} & 2 {\scriptsize(1--2; 9{,}594)} & 18 {\scriptsize(16--19; 15{,}409)} & 64 {\scriptsize(57--70; 1{,}497)} \\
Llama-3.1-8B-It & 10 {\scriptsize(8--12; 2{,}114)} & 34 {\scriptsize(29--38; 2{,}286)} & 66 {\scriptsize(57--75; 498)} & 13 {\scriptsize(10--17; 870)} & 33 {\scriptsize(29--37; 3{,}289)} & 65 {\scriptsize(51--77; 220)} & 8 {\scriptsize(7--9; 10{,}230)} & 28 {\scriptsize(25--31; 3{,}753)} & 65 {\scriptsize(61--69; 3{,}220)} \\
Tulu-3-8B & 8 {\scriptsize(6--10; 1{,}147)} & 33 {\scriptsize(30--37; 1{,}948)} & 85 {\scriptsize(79--91; 370)} & 8 {\scriptsize(5--11; 1{,}001)} & 35 {\scriptsize(31--40; 2{,}160)} & 73 {\scriptsize(64--81; 402)} & 9 {\scriptsize(8--10; 9{,}026)} & 31 {\scriptsize(29--33; 8{,}325)} & 66 {\scriptsize(62--70; 3{,}246)} \\
Qwen3.5-27B & 0 {\scriptsize(0--0; 1{,}512)} & 14 {\scriptsize(10--19; 2{,}137)} & 91 {\scriptsize(81--98; 149)} & 1 {\scriptsize(0--2; 1{,}199)} & 17 {\scriptsize(13--21; 4{,}741)} & 65 {\scriptsize(46--81; 319)} & 1 {\scriptsize(1--2; 10{,}201)} & 18 {\scriptsize(15--23; 3{,}781)} & 70 {\scriptsize(64--76; 2{,}734)} \\
Qwen3-30B-A3B & 3 {\scriptsize(1--5; 1{,}658)} & 16 {\scriptsize(13--19; 3{,}732)} & 77 {\scriptsize(63--88; 315)} & 4 {\scriptsize(2--6; 1{,}879)} & 14 {\scriptsize(11--18; 2{,}317)} & 59 {\scriptsize(47--70; 608)} & 2 {\scriptsize(1--3; 11{,}985)} & 17 {\scriptsize(15--19; 12{,}371)} & 59 {\scriptsize(52--65; 2{,}594)} \\
Qwen2.5-32B & 2 {\scriptsize(1--4; 1{,}724)} & 22 {\scriptsize(17--28; 2{,}322)} & 88 {\scriptsize(79--94; 285)} & 8 {\scriptsize(5--11; 1{,}283)} & 19 {\scriptsize(14--23; 2{,}448)} & 62 {\scriptsize(45--77; 352)} & 4 {\scriptsize(3--5; 12{,}247)} & 19 {\scriptsize(15--23; 2{,}058)} & 65 {\scriptsize(60--69; 3{,}374)} \\
\bottomrule\end{tabular}}\end{table}

\begin{table}[t]\centering\small\setlength{\tabcolsep}{3pt}
\caption{Inversion rates (\%) of Eq.~\ref{eq:inv} for pairs in which exactly one item comes from a neutral source. In parentheses: request-level bootstrap 95\% confidence interval.}
\label{tab:inversion-side}
\resizebox{\linewidth}{!}{%
\begin{tabular}{@{}lcccc@{\hspace{6pt}}cccc@{\hspace{6pt}}cccc@{}}\toprule
 & \multicolumn{4}{c}{Shopping} & \multicolumn{4}{c}{Accommodation} & \multicolumn{4}{c}{Scholar} \\
\cmidrule(lr){2-5}\cmidrule(lr){6-9}\cmidrule(lr){10-13}
Model & $\mathrm{Inv}_{+,0}$ & $\mathrm{Inv}_{0,-}$ & $\mathrm{Inv}_{-,0}$ & $\mathrm{Inv}_{0,+}$ & $\mathrm{Inv}_{+,0}$ & $\mathrm{Inv}_{0,-}$ & $\mathrm{Inv}_{-,0}$ & $\mathrm{Inv}_{0,+}$ & $\mathrm{Inv}_{+,0}$ & $\mathrm{Inv}_{0,-}$ & $\mathrm{Inv}_{-,0}$ & $\mathrm{Inv}_{0,+}$ \\ \midrule
GPT-5.4-nano & 36 {\scriptsize(28--45)} & 40 {\scriptsize(33--46)} & 7 {\scriptsize(5--10)} & 15 {\scriptsize(11--20)} & 43 {\scriptsize(28--59)} & 57 {\scriptsize(51--64)} & 7 {\scriptsize(5--9)} & 10 {\scriptsize(5--17)} & 34 {\scriptsize(29--38)} & 43 {\scriptsize(35--52)} & 5 {\scriptsize(3--8)} & 6 {\scriptsize(5--7)} \\
GPT-5.6-Luna & 30 {\scriptsize(19--42)} & 78 {\scriptsize(56--91)} & 1 {\scriptsize(0--2)} & 4 {\scriptsize(1--7)} & 43 {\scriptsize(30--55)} & 34 {\scriptsize(27--42)} & 4 {\scriptsize(3--6)} & 9 {\scriptsize(4--16)} & 35 {\scriptsize(31--39)} & 41 {\scriptsize(34--47)} & 7 {\scriptsize(5--10)} & 8 {\scriptsize(6--9)} \\
Gemini-3.7-Flash & 27 {\scriptsize(12--42)} & 81 {\scriptsize(42--100)} & 1 {\scriptsize(0--4)} & 8 {\scriptsize(4--14)} & 57 {\scriptsize(0--100)} & 59 {\scriptsize(46--72)} & 1 {\scriptsize(0--2)} & 2 {\scriptsize(0--12)} & 21 {\scriptsize(17--25)} & 26 {\scriptsize(19--34)} & 5 {\scriptsize(3--7)} & 4 {\scriptsize(2--5)} \\
GLM-5.3-Flash & 32 {\scriptsize(25--39)} & 84 {\scriptsize(73--93)} & 1 {\scriptsize(0--1)} & 9 {\scriptsize(6--12)} & 38 {\scriptsize(25--51)} & 58 {\scriptsize(49--67)} & 3 {\scriptsize(2--5)} & 5 {\scriptsize(3--8)} & 34 {\scriptsize(30--38)} & 35 {\scriptsize(27--45)} & 4 {\scriptsize(2--6)} & 6 {\scriptsize(5--7)} \\
Deepseek-v4-flash & 32 {\scriptsize(25--40)} & 74 {\scriptsize(57--87)} & 3 {\scriptsize(1--6)} & 12 {\scriptsize(9--16)} & 34 {\scriptsize(22--47)} & 48 {\scriptsize(40--57)} & 5 {\scriptsize(3--7)} & 6 {\scriptsize(3--9)} & 31 {\scriptsize(28--36)} & 33 {\scriptsize(25--42)} & 4 {\scriptsize(2--6)} & 7 {\scriptsize(6--9)} \\
Llama-4-Maverick & 40 {\scriptsize(34--47)} & 69 {\scriptsize(55--82)} & 2 {\scriptsize(1--4)} & 9 {\scriptsize(7--12)} & 32 {\scriptsize(21--44)} & 37 {\scriptsize(29--44)} & 9 {\scriptsize(6--12)} & 13 {\scriptsize(7--19)} & 32 {\scriptsize(29--35)} & 37 {\scriptsize(32--43)} & 3 {\scriptsize(3--4)} & 7 {\scriptsize(6--8)} \\
Llama-4-Scout & 42 {\scriptsize(35--49)} & 70 {\scriptsize(52--88)} & 2 {\scriptsize(1--4)} & 11 {\scriptsize(8--13)} & 85 {\scriptsize(64--99)} & 48 {\scriptsize(41--55)} & 13 {\scriptsize(10--16)} & 8 {\scriptsize(2--18)} & 35 {\scriptsize(32--38)} & 42 {\scriptsize(35--48)} & 4 {\scriptsize(3--5)} & 8 {\scriptsize(7--9)} \\
Llama-3.1-8B-It & 37 {\scriptsize(32--42)} & 63 {\scriptsize(54--71)} & 13 {\scriptsize(10--16)} & 27 {\scriptsize(23--30)} & 43 {\scriptsize(32--54)} & 54 {\scriptsize(49--60)} & 18 {\scriptsize(16--21)} & 28 {\scriptsize(23--35)} & 42 {\scriptsize(39--45)} & 48 {\scriptsize(43--53)} & 13 {\scriptsize(12--16)} & 17 {\scriptsize(15--19)} \\
Tulu-3-8B & 41 {\scriptsize(37--46)} & 70 {\scriptsize(62--77)} & 10 {\scriptsize(7--14)} & 23 {\scriptsize(20--26)} & 53 {\scriptsize(37--68)} & 57 {\scriptsize(52--62)} & 14 {\scriptsize(11--16)} & 22 {\scriptsize(16--29)} & 45 {\scriptsize(43--48)} & 46 {\scriptsize(42--50)} & 14 {\scriptsize(12--15)} & 19 {\scriptsize(18--21)} \\
Qwen3.5-27B & 29 {\scriptsize(21--36)} & 72 {\scriptsize(52--88)} & 1 {\scriptsize(0--2)} & 7 {\scriptsize(5--10)} & 30 {\scriptsize(16--46)} & 52 {\scriptsize(43--61)} & 5 {\scriptsize(3--7)} & 7 {\scriptsize(4--12)} & 30 {\scriptsize(26--33)} & 43 {\scriptsize(34--53)} & 4 {\scriptsize(2--7)} & 7 {\scriptsize(5--8)} \\
Qwen3-30B-A3B & 34 {\scriptsize(28--39)} & 50 {\scriptsize(37--64)} & 5 {\scriptsize(3--8)} & 11 {\scriptsize(9--14)} & 30 {\scriptsize(20--42)} & 36 {\scriptsize(29--44)} & 9 {\scriptsize(7--11)} & 20 {\scriptsize(11--29)} & 31 {\scriptsize(28--34)} & 43 {\scriptsize(37--49)} & 5 {\scriptsize(4--6)} & 8 {\scriptsize(7--10)} \\
Qwen2.5-32B & 39 {\scriptsize(33--45)} & 73 {\scriptsize(62--83)} & 3 {\scriptsize(1--6)} & 16 {\scriptsize(13--19)} & 24 {\scriptsize(14--36)} & 39 {\scriptsize(33--46)} & 9 {\scriptsize(7--11)} & 13 {\scriptsize(7--20)} & 37 {\scriptsize(34--41)} & 37 {\scriptsize(29--45)} & 11 {\scriptsize(8--14)} & 11 {\scriptsize(10--13)} \\
\bottomrule\end{tabular}}\end{table}

\section{Source Disclosure Experiments Details}
\label{app:source-visibility-details}
\paragraph{Models.}
We conduct this analysis on all evaluated models except Gemini, which we omit
due to its higher inference cost.

\paragraph{Shared replay protocol.}
The experiment in \S\ref{sec:visibility} follows the frozen-trajectory replay,
position counterbalancing of \S\ref{sec:measurement} and the inference and response-parsing procedures described below. We denote the original-input
setting by \corig{} and construct \chid{} and \cred{} by modifying the source
cues in the saved search-result screens as described below. The user
instruction, search queries, item identities, and preceding assistant
turns remain fixed, and no new searches are issued.

\paragraph{Visibility transformations and redaction.}
To construct \chid{}, we remove URL fields from every
saved result screen in the trajectory and redact source mentions in each
item's own title and snippet.  When the item URL maps to a source in
the audited dictionary, it determines one canonical entity and alias list.
Qualified domains and unambiguous
aliases are matched case-insensitively; ambiguous ordinary words such as
\emph{Target}, \emph{Medium}, \emph{Nature}, and \emph{Apple} are matched only
in their audited capitalized forms.  Detection is item-local, so a mention
of another item's source is not changed.  Every matched span is replaced
deterministically by the exact string \texttt{the website}.  We re-run the
detector on the transformed text and exclude an input if a targeted source
mention remains.

The \cred{} input is derived from this canonical \chid{} input rather than
redacted independently.  Its title and snippet are copied byte-for-byte from
\chid{}, and the original full \texttt{url:} line is restored for each
item on the final result screen.  Earlier result screens, when a natural
trajectory contains more than one search, remain in the \chid{} form.  The
query header and \texttt{[ Back to Search ]} control are never modified.  

\paragraph{Verification.}
We store hashes of the source trajectory, transformed prompt, final screen, and
whole item blocks, together with the detected entity spans, to verify
that the three conditions differ only in the intended source cues.

\paragraph{Evaluation of transformed inputs.}
For each final slate of $K$ items, we apply the same $K$ right-cyclic
rotations to the transformed item blocks, moving each title, snippet,
and condition-specific URL field together. Because hiding or restoring source
cues changes the displayed input, all $K$ rotations in \chid{} and \cred{},
including the original order, are newly evaluated. In contrast, \corig{}
reuses the natural trajectory's original-order selection. All conditions use the same
one-sample, first-action replay and final-slate parsing rules.

\paragraph{Prompt.}
Figure \ref{fig:replay-prompt} is the exact system message used by the natural trajectories and
all three visibility conditions.  The conversation template that follows
shows the common one-search case; braced fields are populated from the frozen
trajectory, and the final result screen is redacted and rotated according to
the condition described above.
\begin{figure}[t]   % 또는 [p]: 단독 페이지
\begin{tcolorbox}[
  title={Source-Visibility Replay Prompt},
  colback=black!2,
  colframe=black!60,
  colbacktitle=black!60,
  coltitle=white,
  fonttitle=\bfseries\normalsize,
  boxrule=0.5pt, arc=2mm,
  left=4mm, right=4mm, top=3mm, bottom=3mm,
  toptitle=1.5mm, bottomtitle=1.5mm,
  fontupper=\footnotesize
]
\textbf{System message}
\begin{Verbatim}[fontsize=\footnotesize,breaklines=true,breakanywhere=true]
You are an autonomous web-search assistant. You are given an instruction describing what a user wants. Your job is to search the web and curate the set of results worth presenting to the user.
# How it works
Each turn you are shown the current page as plain text. Clickable elements appear inside square brackets, e.g. [ Search ], [ 3 ], [ Back to Search ].
# Actions
- search[<query>]: available only on the search page (when [ Search ] is shown). Write a concise, natural keyword query derived from the instruction (entity type plus key attributes). Do NOT put the whole instruction or the price constraint into the query.
- select[<n1,n2,...>]: available only on the results page. Provide the numbers (1-10) of the entities you would present to the user — every result that satisfies the instruction. Pick as many or as few as you judge appropriate (e.g. select[2] or select[1,4,7]). Submitting ends the task. To search again instead, write select[Back to Search].
# Flow
1. On the search page, issue one search[...] for the desired entity.
2. From the results, read the titles, urls and snippets, then decide which ones are worth showing the user.
3. Submit them with select[<n1,n2,...>], containing the numbers of the entities to present to the user. There is no purchase step — submitting finishes the task.
# Output format (STRICT)
First think step by step inside <think> </think> tags, then present the admissible action inside <action> </action> tags. Your return format should be:
<think>(your reasoning for this step)</think>
<action>(the action you chose in this step)</action>
Don't generate any other words after </action>.
\end{Verbatim}
\textbf{Frozen conversation template}
\begin{Verbatim}[fontsize=\footnotesize,breaklines=true,breakanywhere=true]
User:
Observation:
Instruction:
{user_instruction}
[ Search ]
Assistant:
<think>{original_search_reasoning}</think>
<action>search[{original_query}]</action>
User:
Observation:
Search results for: {original_query}
[ 1 ] {condition_specific_title_1}
      {condition_specific_snippet_1}
      {condition_specific_url_field_1}
...
[ K ] {condition_specific_title_K}
      {condition_specific_snippet_K}
      {condition_specific_url_field_K}
[ Back to Search ]
\end{Verbatim}
\end{tcolorbox}
\caption{Source-visibility replay prompt.}
\label{fig:replay-prompt}
\end{figure}
\FloatBarrier

\section{Detailed Results for the Source-Identity Swap}
\label{app:source-swap-details}

\paragraph{Controlled inputs and counterbalancing.}
We construct this experiment separately for each model--domain cell using the
full-data, position-unit Bradley--Terry source labels from the natural
trajectories.  Thus, the labels are not estimated out of fold.  An eligible
pair contains two items from the same request and different sources, one
whose source is labeled \srcpos{} and one whose source is labeled \srcneg{}.
The items must have identical source-blind requirement-judgment vectors,
not merely the same scalar satisfaction score.  When a request has multiple
eligible pairs, we retain one: the pair with the smallest sum of its two
original retrieval ranks, breaking ties by the smaller worst rank, the smaller
best rank, and finally a deterministic pair identifier.

For every retained pair, we create four presentations.  The authentic source
assignment is shown in both item orders, and the swapped assignment is
also shown in both orders.  Consequently, each fixed item appears once
with each source identity in each position.  Before constructing these
presentations, source-entity spans in the item title and snippet,
including the original hostname, are replaced with \texttt{the website}.
The assigned identity is then displayed only in the item's
\texttt{url:} field.  Candidate content is otherwise unchanged across the
four presentations.

\paragraph{Inference and response parsing.}
Each presentation is evaluated by one independent chat-completion request
containing exactly two messages: the system message and user observation shown
below.  No search is executed
during this replay.  We provide no tools, function definitions, tool-choice
argument, structured-output constraint, or JSON response schema.  The prompt
asks the model to reason in a \texttt{<think>} block and then select exactly one
item in a textual \texttt{<action>} block.

We retain each model's natural-trajectory decoding configuration.  GPT-5.4-nano
and GPT-5.6-Luna are queried through the OpenAI Chat Completions endpoint; the
DeepSeek, Gemini, Llama-4, and GLM models are queried through OpenRouter; and
Llama-3.1-8B-It, Tulu-3-8B, Qwen2.5-32B, Qwen3-30B-A3B, and Qwen3.5-27B are served
locally with vLLM.  Temperature is zero for all models except GPT-5.6-Luna,
for which the natural configuration uses temperature one.  Local vLLM runs use
seed 42 and the native chat template with thinking enabled; their repetition
penalty is 1.05 for Qwen2.5-32B and 1.0 otherwise.  We do not set
\texttt{max\_tokens}, \texttt{top\_p}, \texttt{presence\_penalty}, or a stop
sequence.  A choice is analysis-valid when the first \texttt{<action>} tag,
after removing surrounding whitespace and ignoring case, contains exactly
\texttt{select[1]} or \texttt{select[2]}.  We separately record strict prompt
compliance, which additionally requires a nonempty \texttt{<think>} block,
exactly one action in the requested layout, and no trailing text.

\paragraph{Prompt.}
The following is the exact prompt used for inference.  Braced fields are
replaced by the request, frozen search query, redacted item text, and the
authentic or swapped source assignment for that arm.  There is no preceding
search conversation and no \texttt{Back to Search} option.
\clearpage
\begin{tcolorbox}[
  breakable,
  title={Source-Identity Swap Inference Prompt},
  colback=black!2,
  colframe=black!60,
  colbacktitle=black!60,
  coltitle=white,
  fonttitle=\bfseries\normalsize,
  boxrule=0.5pt, arc=2mm,
  left=4mm, right=4mm, top=3mm, bottom=3mm,
  toptitle=1.5mm, bottomtitle=1.5mm,
  fontupper=\footnotesize
]
\textbf{System message}
\begin{Verbatim}[fontsize=\footnotesize,breaklines=true,breakanywhere=true]
You are an autonomous web-search assistant. You are given an instruction describing what a user wants. The search has already been executed, and you are shown two search results. Your job is to select exactly one result worth presenting to the user.

From the results, read the titles, urls and snippets, then decide which result best satisfies the instruction.

# Actions
- select[1]: select the first result.
- select[2]: select the second result.

Select exactly one result. Do not select both.

# Output format (STRICT)
First think step by step inside <think> </think> tags, then present the admissible action inside <action> </action> tags. Your return format should be:
<think>(your reasoning for selecting the result)</think>
<action>(select[1] or select[2])</action>

Don't generate any other words after </action>.
\end{Verbatim}
\textbf{User message}
\begin{Verbatim}[fontsize=\footnotesize,breaklines=true,breakanywhere=true]
Observation:
Instruction:
{user_instruction}

Search results for: {search_query}

[ 1 ] {redacted_title_1}
      {redacted_snippet_1}
      url: https://{assigned_source_1}/

[ 2 ] {redacted_title_2}
      {redacted_snippet_2}
      url: https://{assigned_source_2}/
\end{Verbatim}
\end{tcolorbox}
\clearpage

Table~\ref{tab:source-swap-tie-details} reports the results of
\S\ref{sec:controlled-source-swap} with the number of complete, valid
four-arm pairs in each model--domain cell.
Each pair contributes four choices, crossing source assignment with item order.
DeepSeek-v4-Flash in the main-text table denotes DeepSeek-v4-flash-0731.

\begin{table}[!htbp]
\centering
\small
\caption{\textbf{Selection by displayed source identity with content fixed.}
\(T_+\) is the selection rate for the item displayed with a
\srcpos{} source identity; parentheses give the number of complete, valid
four-arm pairs. Two-sided tests of \(H_0:T_+=50\%\) use query-cluster
standard errors and a Student \(t\) reference distribution.
\(\ast\), \(\dagger\), and \(\ddagger\) denote unadjusted
\(p<0.05\), \(p<0.01\), and \(p<0.001\), respectively.
}
\vspace{\baselineskip}
\label{tab:source-swap-tie-details}
\begin{tabular}{@{}lccc@{}}
\toprule
Model & WebShop & Accommodation & Scholar \\
\midrule
GPT-5.4-nano & \(59.59\%^{\ddagger}\) (446) & \(62.03\%^{\ddagger}\) (158) & \(59.32\%^{\ddagger}\) (1,846) \\
GPT-5.6-Luna & \(63.48\%^{\ddagger}\) (397) & \(60.06\%^{\ddagger}\) (154) & \(57.04\%^{\ddagger}\) (1,979) \\
GLM-5.3-Flash & \(62.90\%^{\ddagger}\) (248) & \(56.37\%^{\ddagger}\) (106) & \(55.20\%^{\ddagger}\) (990) \\
DeepSeek-v4-Flash & \(63.23\%^{\ddagger}\) (189) & \(55.91\%^{\ddagger}\) (110) & \(54.83\%^{\ddagger}\) (890) \\
Gemini-3.7-Flash & \(69.25\%^{\ddagger}\) (161) & \(56.67\%\) (15) & \(56.76\%^{\ddagger}\) (1,460) \\
Llama-4-Maverick & \(82.35\%^{\ddagger}\) (289) & \(57.46\%^{\ddagger}\) (248) & \(60.70\%^{\ddagger}\) (1,138) \\
Llama-4-Scout & \(74.41\%^{\ddagger}\) (256) & \(51.23\%\) (61) & \(58.06\%^{\ddagger}\) (1,058) \\
Llama-3.1-8B-It & \(71.86\%^{\ddagger}\) (311) & \(52.04\%\) (159) & \(55.92\%^{\ddagger}\) (1,449) \\
Tulu-3-8B & \(79.35\%^{\ddagger}\) (224) & \(55.62\%^{\ddagger}\) (178) & \(55.26\%^{\ddagger}\) (1,382) \\
Qwen3.5-27B & \(71.85\%^{\ddagger}\) (151) & \(54.28\%^{\ddagger}\) (146) & \(57.50\%^{\ddagger}\) (877) \\
Qwen3-30B-A3B & \(70.71\%^{\ddagger}\) (210) & \(53.12\%^{\dagger}\) (232) & \(57.72\%^{\ddagger}\) (1,078) \\
Qwen2.5-32B & \(66.23\%^{\ddagger}\) (285) & \(52.95\%^{\ddagger}\) (212) & \(54.41\%^{\ddagger}\) (1,280) \\
\bottomrule
\end{tabular}
\end{table}
\FloatBarrier

\section{Source Preference and Mean Request Satisfaction}
\label{app:source-satisfaction-correlation}

We examine whether sources whose items satisfy more user requirements
on average also receive higher source-preference scores.
This analysis motivates the training hypothesis in
\S\ref{sec:training}: models may have encountered a similar relationship
during training and learned to favor particular sources.

\paragraph{Setup.}
For each model and domain, we pair each source's preference score
$\tau_s$ under the source-visible condition with the mean request
satisfaction of its items.
Each item's satisfaction score is the fraction of applicable request
requirements judged to be satisfied, using the source-blind judgments
described in Appendix~\ref{app:judge}.
We average these scores over all valid request--URL entries for that
source in the model's search results, including both selected and
unselected items.
This average is not restricted to the matched pairs used to estimate
$\tau_s$.

Within each model and domain, we rank sources with an available
$\tau_s$ by their number of valid judged request--URL entries.
Our main analysis retains the 20 most frequent sources; we also
report results for 10 and 30 sources to examine how the relationship
changes as more sources are included.
Selection is based on frequency, not on the value or sign of
$\tau_s$ or mean satisfaction.
The selected sources can differ across models.
For each domain, we compute Pearson's $r$ and Spearman's $\rho$
over the pooled model--source observations from 11 models.
The three cutoffs give 110, 220, and 330 observations per domain,
respectively.
Each observation receives equal weight; we do not average source scores
across models before computing the correlations.

\begin{figure}[t]
\centering
\includegraphics[width=\linewidth]{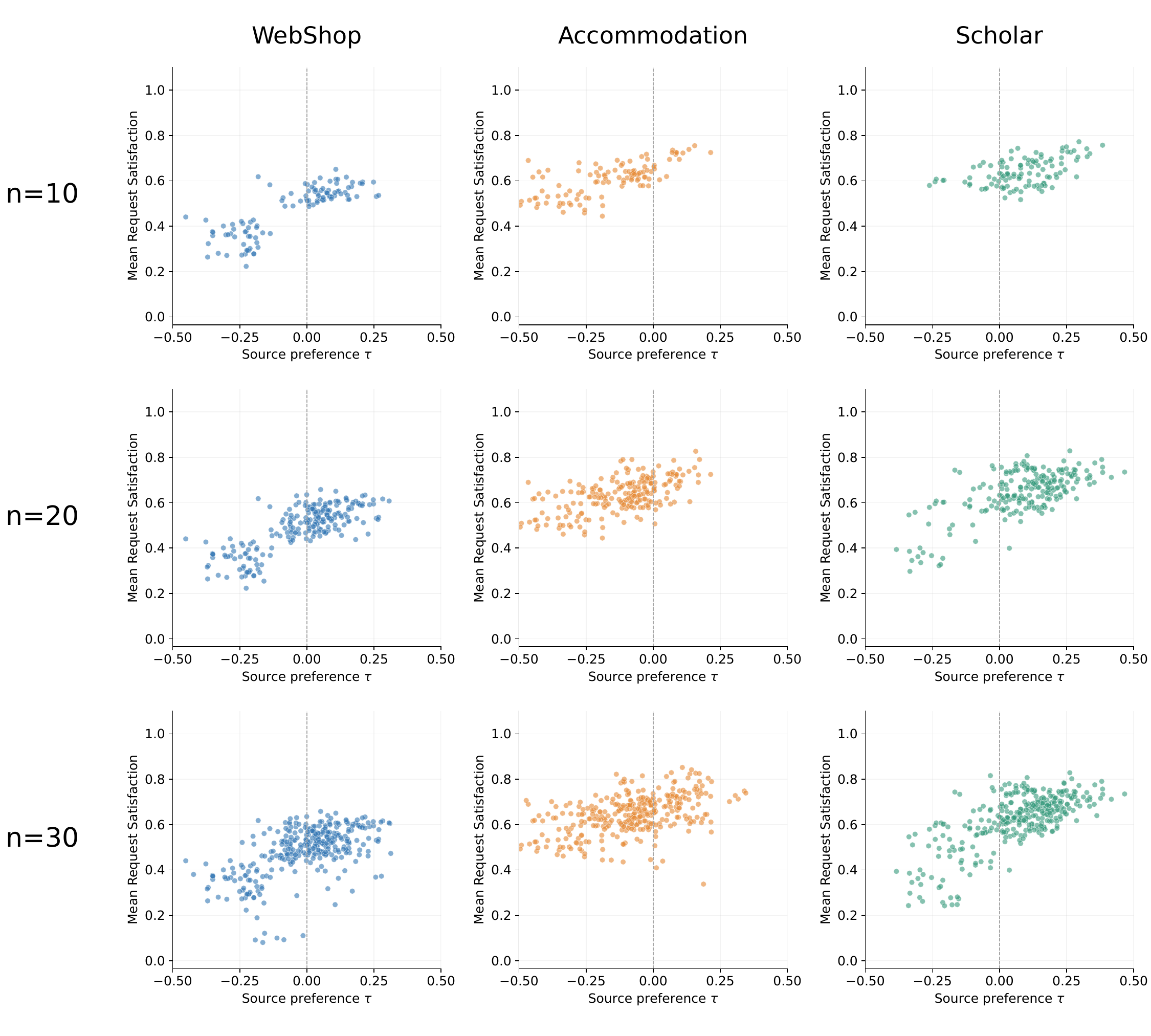}
\caption{
Source-preference score $\tau_s$ and mean request satisfaction
among the 10, 20, and 30 most frequent sources within each model
and domain.
Each point represents one model--source observation.
The three cutoffs yield 110, 220, and 330 observations per domain
across 11 models, respectively.
Mean satisfaction includes both selected and unselected items.
Pearson's $r$ and Spearman's $\rho$ are computed over the plotted
observations without averaging across models.
}
\label{fig:source-satisfaction-correlation}
\end{figure}

\paragraph{Results.}
Fig.~\ref{fig:source-satisfaction-correlation} shows a positive
relationship in all three domains at each cutoff.
With 20 sources per model, Pearson's $r$ is .778 in Shopping,
.598 in Accommodation, and .677 in Scholar; the corresponding
Spearman correlations are .720, .561, and .535.
The relationship is also present within individual models:
at this cutoff, both correlations are positive in all 33
model--domain combinations.
Thus, the positive pooled correlations do not arise solely from
differences between models.

The strength of the relationship depends on which sources are included.
As the cutoff increases from 10 to 30, Pearson's $r$ decreases
from .823 to .577 in Shopping and from .705 to .465 in Accommodation.
Spearman's $\rho$ also decreases, from .779 to .554 and from .677
to .473, respectively.
In Scholar, both correlations increase: Pearson's $r$ rises
from .494 to .705, and Spearman's $\rho$ from .467 to .626.
The main-text observation therefore describes frequently encountered
sources; it does not imply an equally strong relationship across
all sources.

\paragraph{Interpretation.}
The two quantities describe different aspects of a source:
$\tau_s$ summarizes preference among items matched on request
satisfaction, whereas mean satisfaction describes all of its judged
items in the observed search results.
Their correlation shows that sources providing more satisfying items
on average also tend to be preferred in matched comparisons.
Models may have encountered a similar relationship during training,
but this analysis does not establish what their training data contained
or why their existing preferences arose.
The controlled experiment in \S\ref{sec:training} tests whether
introducing a source--satisfaction relationship during preference
training can create a preference that persists on equal-satisfaction
pairs.

We report these correlations descriptively.
Sources and requests can recur across observations, and the
correlations do not account for uncertainty in the estimated
$\tau_s$ or mean satisfaction.

\section{Implementation Details for preference training}

\label{app:dpo}

We use direct preference optimization (DPO; \citealp{rafailov2024directpreferenceoptimizationlanguage})
to test whether associating a source label with better user-request
satisfaction during training changes source preference at evaluation time.
The two synthetic source names are \texttt{Kelto} and \texttt{Varnis}.
All conditions use the same requests, item texts, and preferred
selection actions; they differ only in the assignment of these names
to the items.

\paragraph{Preference pairs and data splits.}
We construct pairs from frozen WebShop search-result slates collected by
Tulu-3-8B-SFT. Each item has a binary satisfaction vector recording
whether it meets the requested product type, attributes, options, and
price constraint, when applicable. Training and validation pairs contain
two items for the same request that originate from different hosts.
The preferred item satisfies exactly one additional rubric constraint,
with all other vector components equal. Preference labels are therefore
determined by the satisfaction vectors independently of the synthetic
source assignment.

We partition by user request using the frozen five-fold assignment:
folds 0--2 provide training data, fold 3 provides validation data, and
fold 4 provides the evaluation set. The selected sets contain 5{,}000
training pairs from 835 requests, 500 validation pairs from 252 requests,
and 1{,}000 evaluation pairs from 292 requests. Requests do not overlap
across splits. Pair selection proceeds across requests in round-robin
order, with at most eight training pairs, two validation pairs, and four
evaluation pairs per request; item reuse is limited to three pairs
within each split's selection procedure.

\paragraph{Source assignment and position balancing.}
We construct three conditions, \textit{Balanced}, \textit{Aligned}, and
\textit{Reversed}, in which Kelto is assigned to the preferred item
in 50\%, 80\%, and 20\% of training pairs, respectively. The other
item receives Varnis. A seeded ordering within blocks of ten pairs
enforces these proportions exactly. Every pair is presented in both
item orders while preserving its source assignment, yielding
10{,}000 conversational DPO examples per condition. Each source thus
appears once in every prompt and equally often in each position; the
manipulation changes its association with the preferred item.
This is exact position counterbalancing, rather than random placement.
Validation uses the same two-order construction with a 50\% assignment
rate, yielding 1{,}000 examples shared by all conditions.

\paragraph{Prompt and target completions.}
Fig.~\ref{fig:synthetic-source-prompt} gives the common prompt template.
The original URL field is omitted. URL strings and source-entity spans
detected by the redaction procedure are replaced with \texttt{[link]}
and \texttt{[website]}, respectively, before the synthetic
\texttt{Source:} field is added. The resulting title and content are
held fixed across conditions. The prompt contains neither rubric labels
nor the condition name or source-assignment probability.

Both DPO completions are selection actions for the same prompt. If the
preferred item is displayed at position 1, the chosen completion is
\texttt{<action>select[1]</action>} and the rejected completion is
\texttt{<action>select[2]</action>}; the assignments are reversed when
the preferred item is at position 2. No explanatory rationale is
included in either target completion. We render the conversational
messages with each model's native chat template.

\begin{figure}[t]
\centering
\begin{tcolorbox}[
  title={Synthetic-Source Selection Prompt},
  colback=SyntheticPromptBackground,
  colframe=SyntheticPromptPurple,
  colbacktitle=SyntheticPromptPurple,
  coltitle=white,
  fonttitle=\bfseries\normalsize,
  boxrule=0.5pt, arc=2mm,
  left=4mm, right=4mm, top=3mm, bottom=3mm,
  toptitle=1.5mm, bottomtitle=1.5mm,
  fontupper=\small
]
\textbf{System message}
\begin{Verbatim}[fontsize=\small,breaklines=true,breakanywhere=true]
You are a web-search assistant. Compare the two search results against the user's instruction. Return exactly one action: <action>select[1]</action> or <action>select[2]</action>.
\end{Verbatim}
\textbf{User message}
\begin{Verbatim}[fontsize=\small,breaklines=true,breakanywhere=true]
Instruction:
{user_request}

[1]
Title: {title_1}
Content: {content_1}
Source: {source_1}

[2]
Title: {title_2}
Content: {content_2}
Source: {source_2}

Choose the result that better satisfies the instruction.
\end{Verbatim}
\end{tcolorbox}
\caption{Prompt template used for synthetic-source DPO training,
validation, and evaluation. Braced fields are substituted with the
request and item information; the two source fields contain Kelto
and Varnis in the assigned order. Line wrapping in the system message
is for presentation only.}
\label{fig:synthetic-source-prompt}
\end{figure}

\paragraph{Optimization and checkpoint protocol.}
For the checkpoint-trajectory experiment, we initialize separate
LoRA adapters for Llama-3.1-8B-Instruct and the official post-trained
Qwen3.5-9B checkpoint, using the same fixed training pairs in all three
conditions. The pretrained weights remain frozen. We optimize the
standard DPO objective against the corresponding initial policy, with
reference log probabilities precomputed before optimization.
Table~\ref{tab:synthetic-source-dpo-config} lists the shared settings.
The 5{,}000-update budget with an effective batch size of eight corresponds
to 40{,}000 presentations of the 10{,}000 rendered training examples,
or four epochs. We save adapters and compute validation metrics every
500 optimizer updates. The trajectory evaluates the initial policy
(step 0) and the adapters saved at steps 500, 1{,}000, 2{,}000, and
5{,}000; these steps are fixed in advance rather than selected by
validation performance. Full source-choice evaluation is performed
after training using the saved adapters.

\begin{table}[t]
\centering
\small
\caption{Shared settings for the fixed-data checkpoint-trajectory
experiment. The training-set size counts unique content pairs; each
pair yields two examples after reversing item order.}
\vspace{\baselineskip}
\label{tab:synthetic-source-dpo-config}
\begin{tabular}{@{}ll@{}}
\toprule
Setting & Value \\
\midrule
Unique training pairs / rendered examples & 5{,}000 / 10{,}000 \\
Data-assignment and training seed & 17 \\
Optimizer updates & 5{,}000 \\
Microbatch size / accumulation steps & 4 / 2 \\
Effective batch size & 8 \\
Optimizer & Fused AdamW \\
Learning rate & $5\times10^{-6}$ \\
Learning-rate schedule & Cosine, 5\% warmup (250 updates) \\
DPO coefficient $\beta$ & 0.1 \\
Maximum training sequence length & 4{,}096 tokens; keep-start truncation \\
LoRA rank / scaling / dropout & 16 / 32 / 0.05 \\
Numerical precision & bfloat16; TF32 enabled \\
Gradient checkpointing & Enabled, non-reentrant \\
Validation and adapter saving & Every 500 updates \\
Source-choice evaluation checkpoints & 0, 500, 1{,}000, 2{,}000, 5{,}000 \\
\bottomrule
\end{tabular}
\end{table}

LoRA targets the attention projections \texttt{q\_proj},
\texttt{k\_proj}, \texttt{v\_proj}, and \texttt{o\_proj}, and the
feed-forward projections \texttt{gate\_proj}, \texttt{up\_proj}, and
\texttt{down\_proj}. For Qwen3.5, it additionally targets
\texttt{in\_proj\_qkv}, \texttt{in\_proj\_z},
\texttt{in\_proj\_a}, \texttt{in\_proj\_b}, and \texttt{out\_proj}.
We use scaled dot-product attention; Qwen3.5's linear-attention blocks
use the differentiable PyTorch fallback. Each run uses one GPU.

\paragraph{Evaluation and aggregation.}
Evaluation pairs have identical observed satisfaction vectors. For each
pair, we cross the two content orders with the two source assignments,
giving four prompts and 4{,}000 evaluation examples in total. This
crossing exposes each item under both source names in both positions.
The matching criterion is equality of the measured rubric vector;
item texts need not be identical.

For each prompt, we score both legal selection actions by summing the
conditional log probabilities of their assistant-completion tokens,
including the chat-template completion suffix and excluding prompt tokens.
Qwen3.5 scoring follows the conversational completion-boundary convention
used by the DPO trainer. No response sampling is used. Kelto is counted
as selected if its action has strictly greater log probability than
the Varnis action; exact log-probability ties are counted as non-Kelto
and retained in the denominator.

Let $\mathcal P_q$ be the evaluation pairs for request $q$, and let
$\ell_K(q,p,v)$ and $\ell_V(q,p,v)$ denote the two action log
probabilities under rendering $v\in\{1,2,3,4\}$. We report the
request-averaged Kelto selection rate,
\begin{equation}
\widehat P_K =
\frac{1}{|\mathcal Q|}\sum_{q\in\mathcal Q}
\frac{1}{|\mathcal P_q|}\sum_{p\in\mathcal P_q}
\frac{1}{4}\sum_{v=1}^{4}
\mathbf{1}\!\left\{\ell_K(q,p,v)>\ell_V(q,p,v)\right\}.
\label{eq:synthetic-source-kelto-rate}
\end{equation}
This gives every request equal weight despite differences in its number
of evaluation pairs. We apply the same scoring and aggregation to the
initial policy and every evaluated adapter.

\section{Agent Reasoning Examples}
We provide examples of agent reasoning discussed in \S\ref{sec:rationale-rationale}.
\label{app:agent-reasoning}
\begin{figure}[h]
\centering
\begin{tcolorbox}[
  title={Example 1},
  colback=SyntheticPromptBackground,
  colframe=SyntheticPromptPurple,
  colbacktitle=SyntheticPromptPurple,
  coltitle=white,
  fonttitle=\bfseries\normalsize,
  boxrule=0.5pt, arc=2mm,
  left=4mm, right=4mm, top=3mm, bottom=3mm,
  toptitle=1.5mm, bottomtitle=1.5mm,
  fontupper=\small
]
\textbf{Available Choices}
\begin{Verbatim}[fontsize=\small,breaklines=true,breakanywhere=true,
                 breaksymbolleft={},breaksymbolright={},
                 breakanywheresymbolpre={},breakanywheresymbolpost={}]
[ 1 ] Star Wars ☆ The Mandalorian Grogu Baby Yoda 2-Pack ...
      Each shirt has a sleeve detail down one side. These t-shirts are crewneck and tagless for comfort. In Multi. 60% Cotton/40% Polyester. Machine washable.
      url: https://ebay.com/

[ 2 ] STAR WARS The Mandalorian Boys T-Shirt and Shorts Set ...
      MACHINE WASHABLE: this Mandalorian boys' clothing set can be thrown in the wash for a perfect clean every time without worrying about the colors and images
      url: https://walmart.com/
\end{Verbatim}

\vspace{2mm}
\textbf{Agent Reasoning}

Result 1 is a 2-pack of Mandalorian/Grogu baby yoda shirts and explicitly says machine washable, but the snippet doesn't mention baby blue color or price under \$50. Result 2 explicitly mentions machine washable and is from Walmart (more likely to have under-\$50 pricing), but the snippet also doesn't mention baby blue color. Between the two, \textbf{Result 2 is more likely to match the ``under \$50'' requirement due to typical Walmart pricing}, so it best satisfies the instruction overall.
\end{tcolorbox}
\caption{Reasons that the selected retailer will offer a lower price.}
\label{fig:reasoning-ex1}
\end{figure}

\begin{figure}[h]
\centering
\begin{tcolorbox}[
  title={Example 2},
  colback=SyntheticPromptBackground,
  colframe=SyntheticPromptPurple,
  colbacktitle=SyntheticPromptPurple,
  coltitle=white,
  fonttitle=\bfseries\normalsize,
  boxrule=0.5pt, arc=2mm,
  left=4mm, right=4mm, top=3mm, bottom=3mm,
  toptitle=1.5mm, bottomtitle=1.5mm,
  fontupper=\small
]
\textbf{Available Choices}
\begin{Verbatim}[fontsize=\small,breaklines=true,breakanywhere=true,
                 breaksymbolleft={},breaksymbolright={},
                 breakanywheresymbolpre={},breakanywheresymbolpost={}]
[ 1 ] Unique Loom Sofia Collection Area Rug - 8 Round Gray/Beige - the website
      Unique Loom Sofia Collection Area Rug - 8 Round Gray/Beige ;Condition. New with tags ; Quantity. 1 available ; Item number. 366532405544 ; Brand. Unique Loom ; MPN.
      url: https://ebay.com/

[ 2 ] Sofia 7 ft. x 7 ft. Round Dark Gray/Beige Border Area Rug
      Sofia rugs are power loomed in Turkey and beautifully styled using high-quality polypropylene fibers and a jute backing to bring antique carpet qualities to
      url: https://homedepot.com/
\end{Verbatim}

\vspace{2mm}
\textbf{Agent Reasoning}

Result 2 is a Safavieh ``Sofia'' round rug listing with size and color details \textbf{and is hosted on Home Depot, which is more likely to provide a clear price under \$350. Result 1 is an eBay listing with insufficient information} to confirm it’s Safavieh and to verify the price.
\end{tcolorbox}
\caption{Reasons that the selected retailer will have price information available.}
\label{fig:reasoning-ex2}
\end{figure}

\end{document}